\documentclass{article}

\PassOptionsToPackage{numbers, compress}{natbib}
\usepackage[preprint]{neurips_2026}

\usepackage[utf8]{inputenc} 
\usepackage[T1]{fontenc}    
\usepackage{hyperref}       
\usepackage{url}            
\usepackage{booktabs}       
\usepackage{amsfonts}       
\usepackage{nicefrac}       
\usepackage{microtype}      
\usepackage{xcolor}         
\usepackage{graphicx} 
\usepackage{caption} 
\usepackage{xcolor} 
\usepackage{caption} 
\usepackage{natbib} 
\usepackage{multirow} 
\usepackage{adjustbox} 
\usepackage{amsmath} 
\usepackage{wrapfig} 
\usepackage{enumitem} 
\usepackage[most]{tcolorbox} 

\newcommand{\ourmethod}{EgoExo-WM}
\newcommand{\ourmethodspace}{EgoExo-WM }
\newcommand{\ouregox}{EgoX-Body}
\newcommand{\ouregoxspace}{EgoX-Body }

\title{\ourmethod:\\Unlocking Exo Video for Ego World Models}

\author{%
  Danny Tran$^{*\:1}$  \hspace{.4in} Roberto Mart\'{i}n-Mart\'{i}n$^{\dagger}$ \hspace{.4in}  Kristen Grauman$^{\dagger}$ \\ \\
    The University of Texas at Austin
}

\newcommand{\projectpage}{\href{https://vision.cs.utexas.edu/projects/EgoExo-WM/}{project page}}

\begin{document}

\maketitle

\newenvironment{alphafootnotes}
  {\par\edef\savedfootnotenumber{\number\value{footnote}}
   \renewcommand{\thefootnote}{\alph{footnote}}
   \setcounter{footnote}{0}}
  {\par\setcounter{footnote}{\savedfootnotenumber}}
\begin{alphafootnotes}
\phantomsection\let\thefootnote\relax\footnotetext{$\dagger$ Indicates Equal advising.}
\end{alphafootnotes}

\begin{abstract}
Egocentric world models present a promising direction for enabling agents to predict and plan, but their performance is constrained by the limited availability of egocentric data and its inherent partial observability of humans' physical actions. In contrast, exocentric video is abundant and reveals body poses well, but lacks direct alignment with an agent’s action space---and is not egocentric. We propose a method to bridge this gap by extracting structured body pose from exocentric video as a representation of action and transforming the exocentric video to egocentric video, informed by 3D human motion. This process unlocks the integration of in-the-wild exocentric data for egocentric world model training. We show that training whole-body action-conditioned egocentric world models with our converted data significantly improves both prediction quality and downstream planning performance, where we infer the sequence of body poses needed to achieve a visual goal state. Our approach paves the way to enlist arbitrary in-the-wild videos for building powerful egocentric world models, furthering applications in robot planning and augmented-reality guidance. \footnote{Project page:~\url{https://vision.cs.utexas.edu/projects/EgoExo-WM/}.}
\end{abstract}

\section{Introduction}
Egocentric world models aim to provide agents with first-person predictive understanding by modeling how future states evolve under embodied action. Such models could serve as a foundation for future systems that anticipate the consequences of human motion, including robots~\citep{gao2026dreamdojo, assran2025v, mur2026v, goswami2025world}, wearable augmented reality agents~\citep{carmigniani2011augmented}, and assistive technologies that support everyday interactions~\citep{dashner2025human}. In any such human-centered egocentric setting, the relevant action space is 3D human motion: egocentric observations change with the actor's movement and through interactions with the physical space, shaping egocentric observations.

Current egocentric world models are restricted to training with egocentric data~\citep{bai2025whole, pallotta2025egocontrol, tu2025playerone, gao2026dreamdojo}. Unfortunately, learning such models from existing egocentric datasets creates a dual bottleneck. First, egocentric video is difficult to collect at scale because it requires people to wear cameras while performing diverse activities across many environments---today's largest egocentric captures~\cite{grauman2022ego4d,damen2018scaling,ma2024nymeria} are still dwarfed by (largely exocentric) internet-scale video resources~\cite{miech2019howto100m,kay2017kinetics,caba2015activitynet,shan2020understanding}. Second, the 3D human motion needed to define the action space and condition the world model is difficult to obtain at scale: egocentric video provides only a very partial view of the body, and hands are frequently occluded during interaction. 

\begin{center}
    \includegraphics[width=0.95\linewidth]{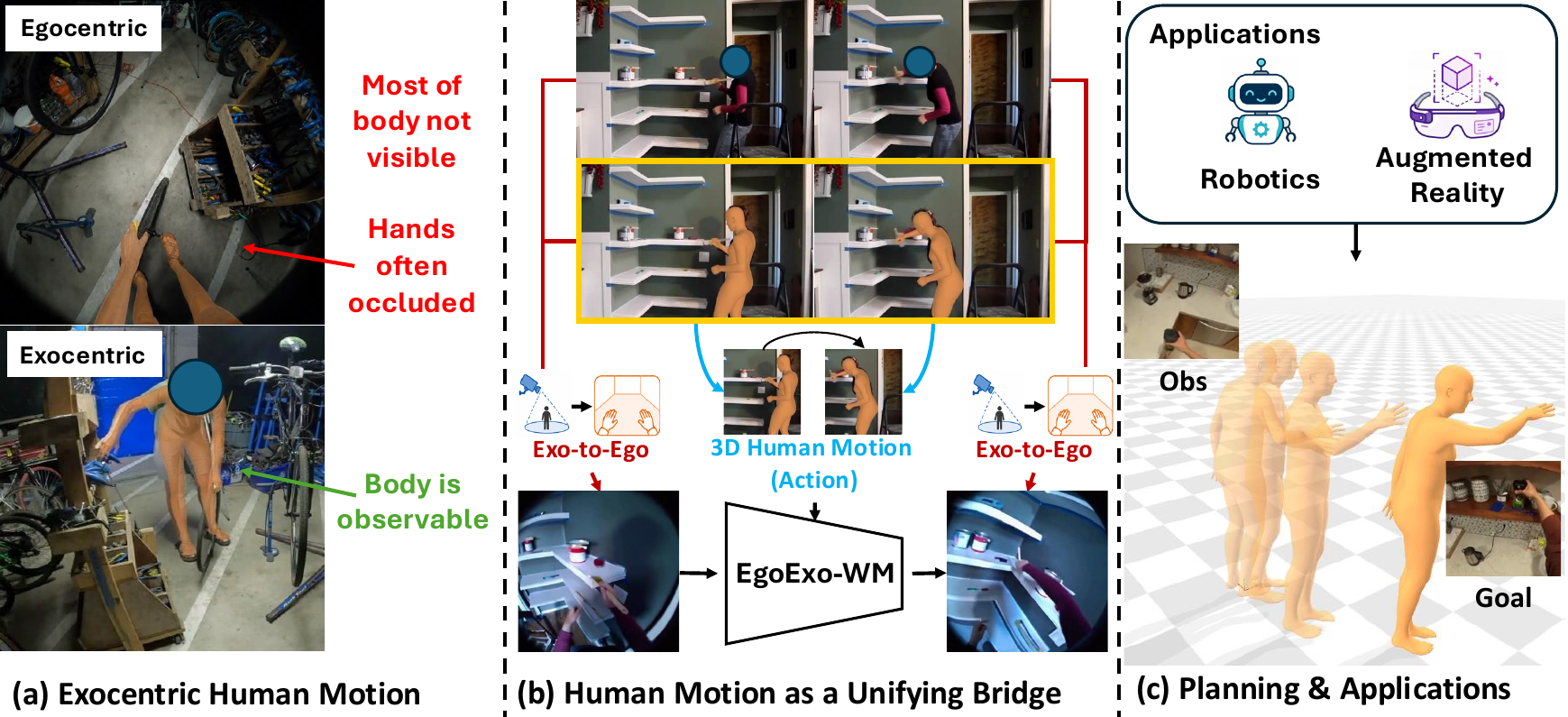}

    \captionsetup{aboveskip=1.25em}
    \begin{minipage}{0.95\textwidth}
        \captionof{figure}{\textbf{Overview.}
    Egocentric video provides an embodied view but often occludes the body and hands, while exocentric video often reveals full-body motion
    (a). \ourmethodspace uses recovered 3D human motion as a bridge for learning egocentric world-models with exocentric video: defining the action
    sequence and guiding exo-to-ego synthesis into action-aligned egocentric
    video (b). The world model enables goal-conditioned planning
    for applications such as robotics and augmented reality by selecting
    actions to reach a target state (c).
    }
        \vspace{-4mm}
        \label{fig:teaser}
    \end{minipage}
\end{center}

We explore a solution to this dual bottleneck that is inspired from human observational learning. Humans can learn about the world not only through direct experience, but also by \emph{observing} others~\citep{bandura1977social, tomasello1993imitative}. We may learn to knead dough from a cooking video, study a friend's brushstrokes while painting, or watch a parent guide thread through the loops of a knitting pattern. From these exocentric observations, we can mentally imagine how the same interactions would unfold through our own eyes: where our hands would move, how objects would shift, resist, or deform under contact, and how each action would reshape the visual scene before us. This ability to transform observed behavior into anticipated first-person experience allows us to learn predictive structure about the world without directly performing every action ourselves~\citep{thirioux2010mental}.  

To translate this ability to AI, we introduce EgoExo-World Model (WM), a framework for leveraging \emph{exocentric} videos to train \emph{egocentric} world models. See Fig.~\ref{fig:teaser}. Exocentric video offers a complementary source of supervision: it exists at internet scale, captures diverse activities and environments, and often reveals the 3D human motion hidden in egocentric views. Our key insight is that 3D human motion can unify exocentric observation, world model learning, and egocentric video synthesis. Specifically, we extract 3D human motion from exocentric video and use it as the action space for world model training and to guide exo-to-ego synthesis. For the latter, our viewpoint conversion module incorporates body motion priors so that synthesized egocentric videos are not only visually plausible, but also aligned with the human action.

Using this approach, we convert diverse exocentric videos from internet sources such as HowTo100M~\citep{miech2019howto100m}, CrossTask~\citep{zhukov2019cross}, and 100 Days of Hands~\citep{shan2020understanding} into paired egocentric video with 3D human body motion sequences. We combine this converted data with existing egocentric data to train a human action-conditioned egocentric world model. Given a history of egocentric observations and a 3D human motion, our world model predicts the future egocentric visual state. By grounding viewpoint conversion and world model learning in 3D human motion, \ourmethodspace transforms exocentric videos into action-conditioned training data for egocentric world models.

Evaluating on three diverse egocentric datasets (Home Action Genome~\citep{rai2021home}, LEMMA~\citep{jia2020lemma}, and Ego-Exo4D~\citep{grauman2024ego}), we demonstrate that \ourmethodspace consistently outperforms SOTA egocentric world-model baselines in future rollout predictions. Beyond prediction, \ourmethodspace enables stronger downstream planning: given contextual egocentric observations and a target visual goal state, the world model can select the sequence of 3D human motions that best drives the agent toward the desired visual outcome---that is, how the person would need to move within the scene and/or manipulate objects to achieve a target state (Fig.~\ref{fig:teaser}(c)). 
Together, these results show \ourmethodspace unlocks exocentric video as a scalable source of visual experience and action supervision for egocentric world modeling.

\section{Related Work}
\label{sec:rw}

\paragraph{World Models.} 
World Models are predictive models that enable agents to predict future observations and reason about the consequences of actions, making them a promising paradigm for embodied intelligence. They have been used for generating synthetic training data~\citep{jang2025dreamgen, alhaija2025cosmos}, improving policies~\citep{quevedo2025worldgym, ebert2018visual, hafner2020mastering}, and learning action-conditioned world representations for planning and control~\citep{gao2026dreamdojo, assran2025v, bar2025navigation, hirose2019vunet, ha2018world, guptamaskvit, wang2026lifting}. Recent work explores egocentric video due to its alignment with the first-person perceptual inputs and onboard sensors used for control~\citep{grauman2022ego4d, zheng2026egoscale, kareer2025egomimic, hao2026egosim, bagchi2026walk, li2026egocentric}. Building on this motivation, recent \emph{egocentric world models} learn predictive dynamics from wearable camera videos, often conditioning on explicit body, hand, or head motion to capture interaction dynamics relevant to humanoid control and dexterous manipulation~\citep{tu2025playerone, bai2025whole, pallotta2025egocontrol, goswami2025world, gao2026lome, hao2026egosim, li2026egocentric}. However, these approaches remain limited by the scarcity of egocentric data with accurate, low-level action annotations---and body pose visibility is inherently restricted in egocentric views. Some works avoid explicit actions by learning latent actions~\citep{gao2026dreamdojo} or conditioning on text descriptions~\citep{yang2023unisim}, but these representations are difficult to interpret or too coarse to capture the fine-grained kinematics needed for embodied control. Thus, a key challenge remains: how to scale egocentric world modeling?

To address this challenge, we scale egocentric world modeling with abundant exocentric video while preserving physically grounded supervision. Our model extracts actions from 3D human pose, providing a richer, more interpretable representation of agent dynamics than latent or low-dimensional actions, and introduces an auxiliary wrist-position consistency loss to ground the learned dynamics in the agent's pose rather than only future visual appearance. Most importantly, we show that transforming exo data to ego unlocks an existing resource for WM training, converting third-person observations into first-person training data.

\vspace*{-0.1in}
\paragraph{Exo-to-Ego Generation.}
Exocentric-to-egocentric view translation synthesizes egocentric observations from exocentric videos, but is challenging due to large viewpoint gaps, occlusions, and ambiguity in human motion. Prior work addresses this with stronger supervision, such as the first egocentric frame~\citep{xu2025egoexo} or synchronized multi-view exocentric videos~\citep{liu2024exocentric}, and with geometric and semantic grounding for cross-view consistency~\citep{cheng20244diff, kang2025egox, barron2021mip, barron2022mip, mildenhall2021nerf, chan2023generative, pumarola2021d, chen2023single, wang2026grounded}. However, these methods emphasize scene-level alignment; methods that transfer human information typically rely on egocentric hand priors~\citep{luo2024put, xu2025egoexo} rather than full-body dynamics. EgoX~\citep{kang2025egox} is a strong framework combining 4D reconstruction~\citep{huang2025vipe}, video generation priors~\citep{wan2025wan}, and VLM-based conditioning~\citep{achiam2023gpt}, but its lack of explicit human prior can lead to action misalignment, including incorrect hand-object interactions, swapped hands, or hallucinated arm motion, as shown in Figure~\ref{fig:egox_body_qualitative}. Our \ouregoxspace builds on EgoX by adding full-body human priors, which provide kinematic cues for projecting hands into the egocentric view consistently with the estimated body pose. We use this conversion step to unlock exocentric videos as training data for egocentric world models. 
Our \ouregoxspace improves downstream world-model training over EgoX (Sec.~\ref{sec:rollout_eval} and Sec.~\ref{sec:wm_planning_results}), demonstrating the value of full-body human priors for exo-to-ego conversion. Our broader contribution is a framework that links exo-to-ego conversion and world-model training through shared 3D human motion, while accommodating EgoX, EgoX-Body and alternative conversion methods~\citep{cheng20244diff, luo2024put, wang2026grounded}.

\vspace*{-0.1in}
\paragraph{Human Pose Estimation.}  
Human pose estimation aims to recover 2D or 3D human structure from images or videos, including body and hand motion~\citep{dwivedi2024tokenhmr, goel2023humans, shin2024wham, ye2023decoupling, xu2022vitpose, pavlakos2024reconstructing, zeng2026stablehand}. These methods either estimate skeletons directly~\citep{lin2014microsoft} or fit parametric models such as SMPL~\citep{loper2023smpl, pavlakos2019expressive}, MANO~\citep{romero2022embodied}, and MHR~\citep{ferguson2025mhr}. Recent work has made substantial progress in improving the accuracy, robustness, and temporal consistency of these predictions across diverse scenes and motions~\citep{gao2025sam, yang2026sam, cai2023smpler, yin2025smplest}. By converting raw visual observations into compact, semantically meaningful structure, pose provides a human prior for downstream tasks such as motion prediction~\citep{ashutosh2025fiction, patel2025uniegomotion, chen2023executing} and robot learning~\citep{li2024okami, bahety2025safemimic, singh2025deep}.
We leverage the human pose present in exocentric videos as a prior in two ways: to guide ego conversion and to serve as the action space for our world model.

\section{\ourmethodspace}
\label{sec:method}

In this section we introduce \ourmethod. We start by presenting our world model (Sec. \ref{sec:wm_training}), followed by our method for converting exocentric videos to egocentric videos (Sec. \ref{sec:exo2ego}). Finally, we describe how we can plan with \ourmethodspace by ranking action sequences (Sec. \ref{sec:wm_planning}). 

\subsection{World Model Training}
\label{sec:wm_training}

\begin{figure}[t]
    \centering
    \includegraphics[width=0.9\linewidth]{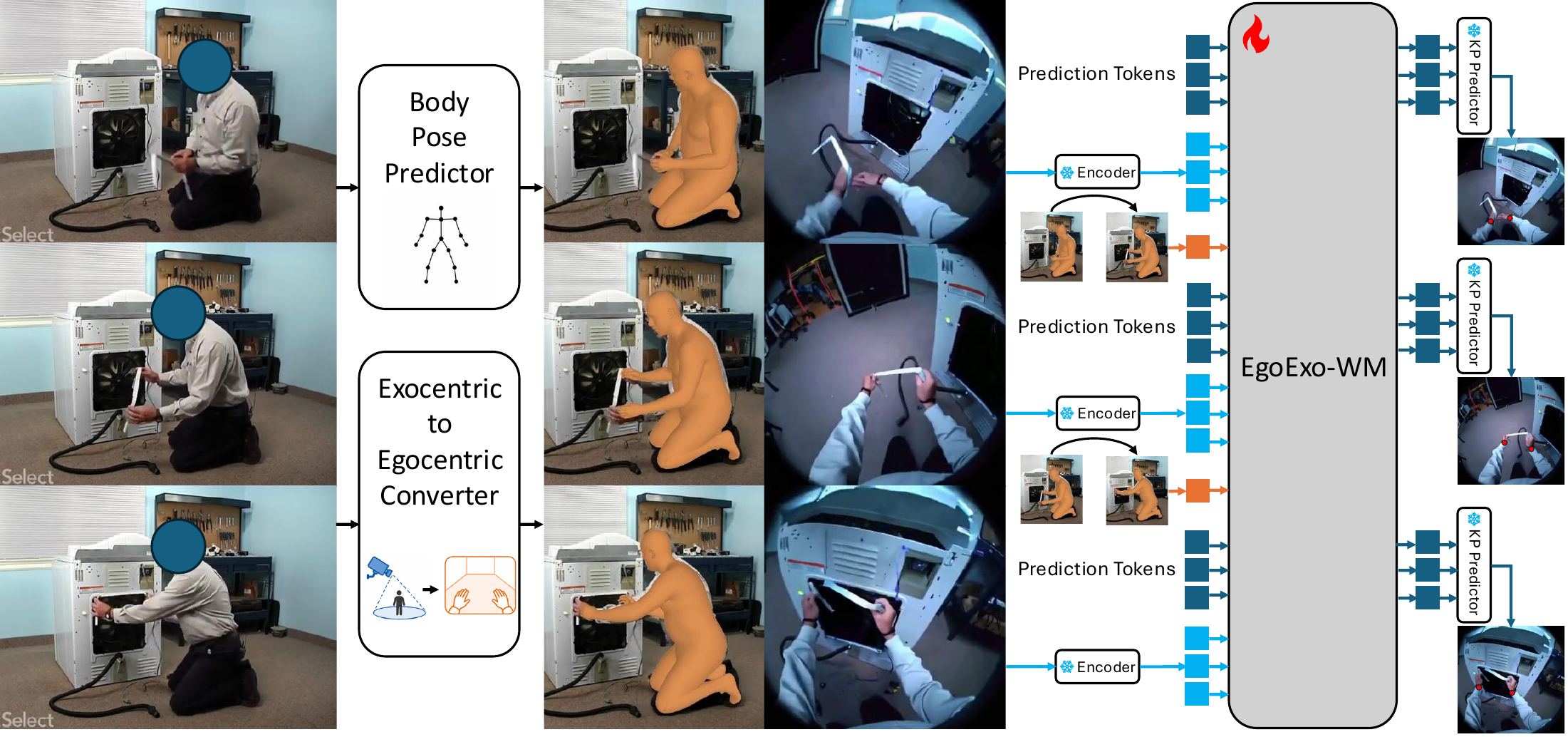}
    \caption{\textbf{World Model Training.} \ourmethodspace unlocks exocentric video for egocentric world model training. Given an exocentric video, we recover 3D human motion and use it alongside the original video to ground exo-to-ego conversion. The 3D human motion becomes our actions and the converted video becomes the egocentric observation. We then train \ourmethodspace autoregressively with teacher forcing. We apply a lightweight wrist keypoint predictor on predicted latents and use a wrist keypoint consistency loss to preserve relevant pose cues.}
    \label{fig:wm_training}
    \vspace{-1.5em}
\end{figure}

We formulate \ourmethodspace as a human motion conditioned egocentric world model. Given a history of egocentric states and a 3D human motion, the model learns to predict how the egocentric state evolves after executing that action. 

Let $\mathbf{x}_t$ denote the egocentric observation at time $t$, and let $\mathbf{a}_t$ denote the 3D human motion action that moves the agent from time $t$ to $t+1$. We represent $\mathbf{a}_t$ as a structured vector that captures both global body motion and local joint articulation. Specifically, $\mathbf{a}_t\in\mathbb{R}^{d_a}$ consists of the change in root translation and the relative rotations of $22$ body joints, with each joint rotation parameterized by Euler angles, and thus $d_a = 3 + 22 \times 3 = 69$.

We train the world model in a visual latent space. Pixel-level prediction is expensive and can overemphasize low-level appearance details, whereas latent prediction provides a more compact target that preserves semantic and geometric information relevant for future egocentric observations. Specifically, we use a pretrained DINOv3-L~\citep{simeoni2025dinov3} encoder $E$ to map each observation $\mathbf{x}_t$ to a visual latent $\mathbf{z}_t = E(\mathbf{x}_t)$. Given a fixed-length context window of $H$ past visual latents $\mathbf{z}_{t-H+1:t}$ and the next 3D human motion action $\mathbf{a}_t$, the world model $f_\theta$ predicts the next visual latent,
\[
    \hat{\mathbf{z}}_{t+1}
    =
    f_\theta(\mathbf{z}_{t-H+1:t}, \mathbf{a}_t).
\]

We supervise this prediction by encoding the future egocentric observation $\mathbf{x}_{t+1}$ with the same visual encoder and regressing to its latent representation $\mathbf{z}_{t+1}=E(\mathbf{x}_{t+1})$ using $\mathcal{L}_{\text{latent}}=\|\hat{\mathbf{z}}_{t+1}-\mathbf{z}_{t+1}\|_2^2$.

Because human pose is central to grounding egocentric dynamics, we encourage the predicted future state to preserve relevant pose cues by adding an auxiliary wrist-position consistency objective.
In our implementation, a lightweight head $h_\phi$ predicts a wrist heatmap $\hat{\mathbf{V}}_{t+1} = h_\phi(\hat{\mathbf{z}}_{t+1})$ from the predicted latent state, which is supervised using wrist pseudo-labels $\mathbf{V}_{t+1}$ extracted from the corresponding training frame. The full training objective is thus
\vspace{-0.1em}
\[
    \mathcal{L}
    =
    \underbrace{\|\hat{\mathbf{z}}_{t+1} - \mathbf{z}_{t+1}\|_2^2}_{\mathcal{L}_{\text{latent}}}
    +
    \lambda
    \underbrace{\|\hat{\mathbf{V}}_{t+1} - \mathbf{V}_{t+1}\|_2^2}_{\mathcal{L}_{\text{wrist}}}.
\]
where $\lambda$ is used to scale the wrist position consistency loss; we find $\lambda = 1$ works well.

This formulation enables joint training on real egocentric and converted exocentric videos in a shared observation-action format. Real egocentric datasets require external sensing for full-body motion, such as the motion-capture suit used in Nymeria~\citep{ma2024nymeria}. For converted exocentric videos, we estimate 3D human motion directly from the exocentric view and pair it with synthesized egocentric observations. Critically, synthesized egocentric videos complement real egocentric videos with a broader range of activities, scenes, and interactions. Concretely, this means we can take a specialized dataset like Nymeria and join it with in-the-wild, broadly available internet video like HowTo100M~\cite{miech2019howto100m} or 100 Days of Hands~\cite{shan2020understanding}. Figure~\ref{fig:wm_training} overviews how we incorporate synthesized egocentric videos. 

\subsection{Exo-to-Ego Data Generation}
\label{sec:exo2ego}
Synthesizing egocentric video from an exocentric view is inherently underdetermined, as viewpoint differences and occlusions obscure the spatial relationships for human-object interactions. To address this, we develop \ouregox, which grounds generation in explicit human kinematic structure to preserve these interactions across views.

Building on Ego-X~\citep{kang2025egox}, \ouregoxspace uses 3D human motion extracted by SAM-Body4D~\citep{gao2025sam} (we provide sensitivity analysis for the pose estimator in Appendix~\ref{sec:pose_estimator_sensitivity})
to provide kinematic conditioning at two complementary levels. On the exocentric side, while the input video contains rich motion data, it is often entangled with viewpoint and background. We introduce kinematic conditioning by overlaying skeletons (we use the MHR representation~\citep{MHR:2025}) onto the conditioning frames. This provides a structured, canonical representation of behavior that grounds the synthesis in the actor's physical movement. 
On the egocentric side, we introduce egocentric hand kinematic conditioning to capture human-object interactions. We use a hand-skeleton overlay to guide consistent hand motion and interaction-relevant cues. We ablate both conditioning components in Appendix~\ref{sec:component_ablation},
showing that both reduce conversion failures, with the egocentric hand prior having a larger impact.

\begin{figure}[t]
    \centering
    \includegraphics[width=\linewidth]{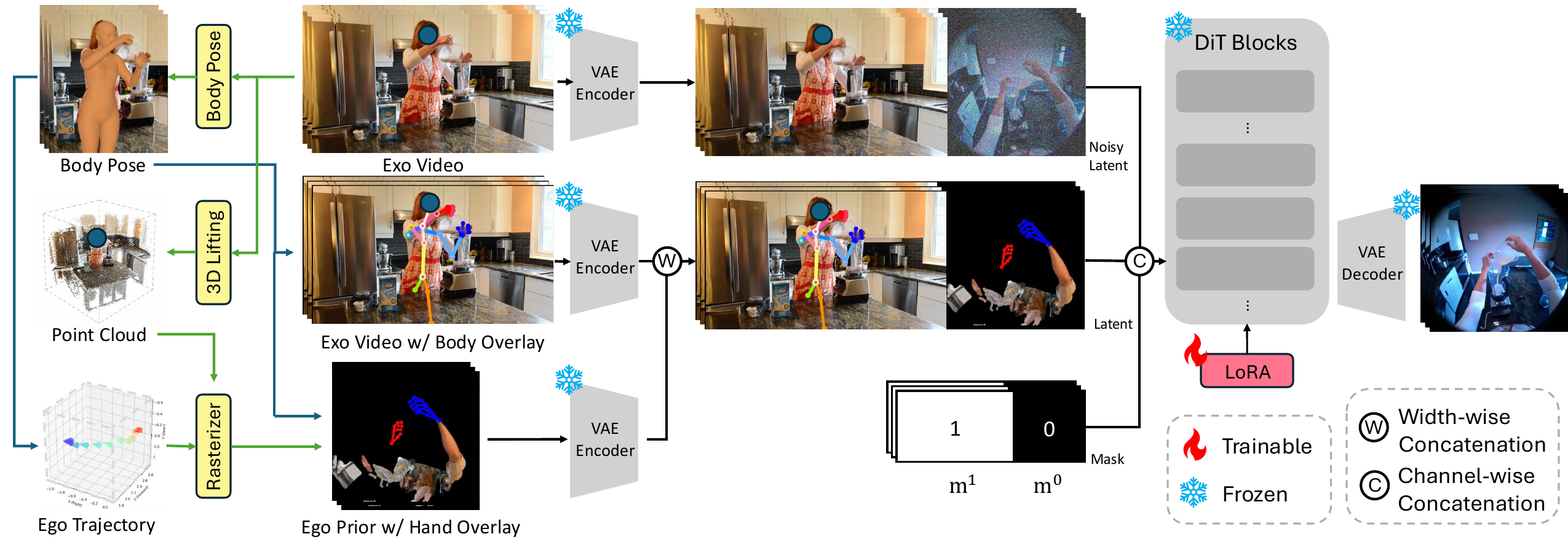}
    \caption{\textbf{\ouregoxspace Inference Overview.} From exocentric videos, we extract body pose and lift the scene into a 3D point cloud. We overlay the body skeleton onto the exocentric video and use the same pose and geometry to render an egocentric prior with predicted hand locations. We form two latent inputs: (1) the clean exocentric latent concatenated with noise, and (2) the body-overlaid exocentric latent concatenated with the egocentric hand prior. These are combined channel-wise with a mask and fed into a video diffusion model to generate egocentric video. Incorporating these human kinematic priors produces outputs that are better grounded in human actions (Fig.~\ref{fig:egox_body_qualitative}).
    }
    \label{fig:egox_body_overview}
    \vspace{-1.5em}
\end{figure}

\begin{wrapfigure}{R}{0.45\textwidth}
    \centering
    \vspace{-7mm}
    \includegraphics[width=\linewidth]{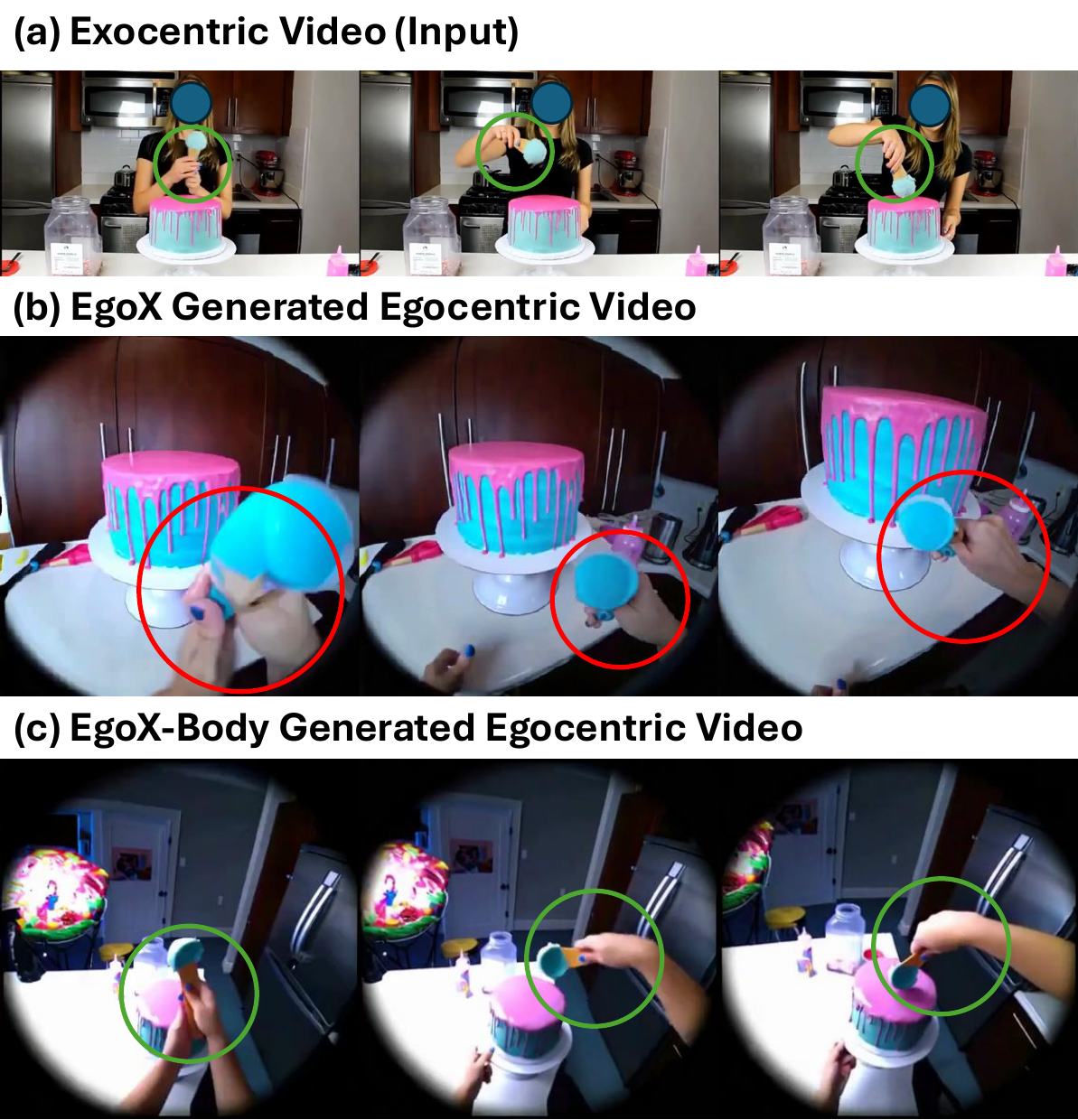}
    \vspace{-4mm}
    \caption{
        \textbf{EgoX-Body Qualitative Comparison.} EgoX-Body better grounds generated egocentric video in human motion and interaction structure. Generated videos misaligned with human motion in EgoX introduce noisy world model training data, leading to inaccurate rollouts and weaker planning (See Sec.~\ref{sec:rollout_eval} and Sec.~\ref{sec:wm_planning_results}).
    }
    \label{fig:egox_body_qualitative}
    \vspace{-8mm}
\end{wrapfigure}
During training, we define hand kinematics using ground-truth poses extracted with HaMeR~\citep{pavlakos2024reconstructing}. 
At inference time, we reconstruct the hand kinematics directly from the exocentric input by projecting the 3D hand positions from the estimated body pose into this rendered view, creating a conditioning signal that aligns the generated egocentric video with the observed human motion. This dual-prior approach ensures that \ouregoxspace generates videos that are not only visually plausible but faithfully represent the underlying human action. For example, we found EgoX-Body better preserves which hand is actually engaged in object manipulation as well as which side of the body the action takes place. Figure~\ref{fig:egox_body_overview} illustrates the inference pipeline. We evaluate \ouregoxspace against EgoX quantitatively for training in Sec.~\ref{sec:rollout_eval} and Sec.~\ref{sec:wm_planning_results} and qualitatively in Figure~\ref{fig:egox_body_qualitative}. We further analyze conversion robustness across varying viewpoints, body occlusion levels, and action complexity in Appendix~\ref{sec:egox_analysis},
failure cases in Appendix~\ref{app:failure-cases},
and additional qualitative examples in Appendix~\ref{sec:egox_more_qual} and on our \projectpage{}.

Importantly, EgoX-Body requires paired ego-exo data only for one-time converter training. Once frozen, it transforms unpaired exocentric videos into egocentric training data, enabling EgoExo-WM to leverage a variety of exocentric datasets (details below) without any further paired data. Thus, the paired-data cost is amortized across large-scale world-model training and future datasets.



\subsection{Planning with \ourmethod}
\label{sec:wm_planning}
Having introduced our world model and the viewpoint conversion model, we now explain how it can be used for planning.  
A key advantage of an action-conditioned egocentric world model is that it can be used not only to predict future observations, but also to choose actions that make progress toward a desired future state. For example, a robot could be given the textual goal 'load my laundry' along with a \emph{visual goal image} of the laundry in the washing machine and subsequently plan a sequence of actions that reach that goal~\citep{ebert2018visual, assran2025v, bar2025navigation, hirose2019deep, ye2026world}. Similarly, in an augmented reality coaching setting, a visual goal could depict a soccer ball reaching the top corner of the soccer goal net. Given the user's current observation, the planner could determine a human motion sequence that is predicted to achieve this outcome, and then provide guidance on how to position the support foot, align the body, strike the ball, and follow through from the user's current position.

We instantiate this idea with model-predictive control (MPC)-style planning over candidate human motion sequences. Given the current egocentric observation and visual goal, we roll out candidate action sequences with our world model and select the one whose predicted future state is closest to the goal. 

At each planning step, we evaluate $N$ candidate action sequences $\{a_{t:t+H-1}^{(i)}\}_{i=1}^{N}$ over horizon $H$ by autoregressively rolling out our world model in latent space, $z_{t+1:t+H}^{(i)} = f_\theta(z_t, a_{t:t+H-1}^{(i)})$, where $z_t$ is the current latent observation. We assume access to a desired goal image, which in practice would be provided by a user-captured image, selected from a demonstration video, or generated from a language instruction describing the desired outcome. We encode the goal image as a target latent $z_g$ and score each candidate sequence by its closest predicted final rollout state to the goal, $\mathcal{C}^{(i)} = \| z_{t+H}^{(i)} - z_g \|_2^2$. We then select the candidate action sequence with the lowest predicted cost:
\[
\hat{i}
=
\arg\min_{i \in \{1,\dots,N\}}
\mathcal{C}^{(i)},
\qquad
a_{t:t+H-1}^{*}
=
a_{t:t+H-1}^{(\hat{i})}.
\]
Planning success therefore depends on the accuracy of \ourmethodspace's predicted rollouts.

\section{Experiments and Results}
\label{sec:exp_and_res}
We describe the setup then provide results.

\subsection{Experimental Setup}

\smallskip
\noindent\textbf{Training Data.}
Our training data consists of (1) real egocentric video and (2) synthetic egocentric data generated from exocentric sources. As \textbf{egocentric training data,} we use Nymeria~\citep{ma2024nymeria}. Nymeria provides egocentric video synchronized with full-body motion capture, recorded in real-world environments such as homes and outdoor parking lots using an Xsens system~\citep{movella_mvn}. We follow the preprocessing steps of PEVA~\citep{bai2025whole}, resulting in $\sim$200 hours of training data.

Our approach further injects converted \textbf{exocentric training data} from multiple large-scale video datasets. We select clips with visible heads and bodies performing actions (details in Appendix~\ref{sec:exo_filtering}),
then convert them to egocentric views using our pipeline in Section~\ref{sec:exo2ego} and Figure~\ref{fig:egox_body_overview}. Our exocentric sources are: (1) HowTo100M~\citep{miech2019howto100m}, a large-scale instructional video dataset, from which we sample $\sim$5 hours from the Food and Entertainment category; (2) CrossTask~\citep{zhukov2019cross}, a task-oriented instructional video dataset, from which we sample $\sim$1 hour; and (3) 100 Days of Hands~\citep{shan2020understanding}, a large collection of everyday hand-object interaction videos, from which we sample $\sim$4 hours. 
We convert a diverse $\sim$10-hour subset drawn from these sources, balancing coverage with the computational cost of video generation. Exo-to-ego conversion is performed offline, taking approximately $3.9$ minutes per $49$-frame clip on a GH200 GPU (details in Appendix~\ref{sec:egox_body_inference_runtime}),
with potential for further acceleration using recent efficient video diffusion methods~\citep{zhang2025turbodiffusion, lyu2025fastercache, nie2026transition, gao2026dreamdojo}. Our framework can incorporate additional converted exocentric clips as compute resources permit.

Notably, despite being trained on a much smaller paired ego-exo dataset, \ouregoxspace generalizes to novel interactions in familiar settings (e.g., kneading dough, whisking, and cake decoration) and activities in unseen settings (e.g., car maintenance, woodworking, and gardening), unlocking previously unpaired exocentric videos for world-model training (see Appendix~\ref{sec:training_data_stats}
for more examples and statistics).

\smallskip
\noindent\textbf{Evaluation Data.}
We evaluate on four diverse held-out datasets to assess both egocentric latent state prediction and downstream planning. Unlike prior methods evaluated primarily on training-domain datasets~\citep{bai2025whole, pallotta2025egocontrol}, our evaluation spans broader variation in activities, environments, and action sequences. We use: (1) Home Action Genome (HOMAGE)~\citep{rai2021home}, paired egocentric/exocentric videos of household activities such as cooking, cleaning, laundry, and grooming; (2) LEMMA~\citep{jia2020lemma}, paired egocentric/exocentric videos of daily activities such as watering plants, making cereal, and sweeping. For HOMAGE and LEMMA, we extract full-body poses with SAM-Body4D and sample $150$ clips, using single person videos; (3) Ego-Exo4D~\citep{grauman2024ego}, a large-scale paired egocentric/exocentric dataset for which we use provided UniEgoMotion poses~\citep{patel2025uniegomotion} and sample $200$ clips each from the Bike and Cooking subsets.

\begin{table*}[t]
    \centering
    \scriptsize
    \setlength{\tabcolsep}{2.5pt}
    \renewcommand{\arraystretch}{1.12}
    \begin{adjustbox}{max width=\textwidth}
    \begin{tabular}{
        lc
        *{3}{cccccc@{\hspace{3pt}}}
        cccccc
    }
        \toprule
        \multirow{3}{*}{Model}
        & \multirow{3}{*}{TFLOPs $\downarrow$}
        & \multicolumn{6}{c}{\textbf{LEMMA}}
        & \multicolumn{6}{c}{\textbf{HOMAGE}}
        & \multicolumn{6}{c}{\textbf{Ego-Exo4D-Bike}}
        & \multicolumn{6}{c}{\textbf{Ego-Exo4D-Cooking}} \\
        \cmidrule(lr){3-8}
        \cmidrule(lr){9-14}
        \cmidrule(lr){15-20}
        \cmidrule(lr){21-26}

        & & \multicolumn{2}{c}{DINO $\downarrow$}
        & \multicolumn{2}{c}{PE $\downarrow$}
        & \multicolumn{2}{c}{Wrist Acc $\uparrow$}
        & \multicolumn{2}{c}{DINO $\downarrow$}
        & \multicolumn{2}{c}{PE $\downarrow$}
        & \multicolumn{2}{c}{Wrist Acc $\uparrow$}
        & \multicolumn{2}{c}{DINO $\downarrow$}
        & \multicolumn{2}{c}{PE $\downarrow$}
        & \multicolumn{2}{c}{Wrist Acc $\uparrow$}
        & \multicolumn{2}{c}{DINO $\downarrow$}
        & \multicolumn{2}{c}{PE $\downarrow$}
        & \multicolumn{2}{c}{Wrist Acc $\uparrow$} \\

        \cmidrule(lr){3-4} \cmidrule(lr){5-6} \cmidrule(lr){7-8}
        \cmidrule(lr){9-10} \cmidrule(lr){11-12} \cmidrule(lr){13-14}
        \cmidrule(lr){15-16} \cmidrule(lr){17-18} \cmidrule(lr){19-20}
        \cmidrule(lr){21-22} \cmidrule(lr){23-24} \cmidrule(lr){25-26}

        & & 2s & Avg & 2s & Avg & 2s & Avg
        & 2s & Avg & 2s & Avg & 2s & Avg
        & 2s & Avg & 2s & Avg & 2s & Avg
        & 2s & Avg & 2s & Avg & 2s & Avg \\
        \midrule

        PEVA$^*$
        & 1304
        & .088 & .072 & .674 & .601 & .321 & .426
        & .091 & .077 & .684 & .630 & .270 & .338
        & .076 & .065 & .648 & .597 & .344 & .407
        & .083 & .070 & .658 & .602 & .294 & .354 \\

        PEVA
        & 1304
        & .115 & .107 & .745 & .704 & .380 & .445
        & .116 & .109 & .741 & .709 & .303 & .397
        & .110 & .100 & .711 & .674 & .298 & .426
        & .106 & .098 & .720 & .684 & .228 & .304 \\

        EgoControl$^*$
        & 356
        & .093 & .079 & .702 & .629 & .352 & .463
        & .100 & .087 & .706 & .651 & .377 & .462
        & .085 & .073 & .667 & .614 & .387 & .526
        & .092 & .079 & .691 & .630 & .246 & .382 \\

        \midrule

        Ego-WM
        & 6.1
        & .068 & .055 & .646 & .567 & .433 & .527
        & .069 & .058 & .657 & .602 & .313 & .396
        & .050 & .042 & \textbf{.602} & .546 & .468 & .561
        & .063 & .053 & \textbf{.640} & .585 & \textbf{.460} & .459 \\

        Naive EgoExo-WM
        & 6.1
        & .064 & .049 & .657 & .555 & .439 & .561
        & .065 & .053 & .651 & .578 & .347 & .447
        & \textbf{.048} & \textbf{.040} & .605 & .549 & .382 & .525
        & \textbf{.062} & \textbf{.052} & \textbf{.640} & .584 & .368 & .443 \\

        Naive EgoExo-WM$^\dagger$
        & 6.1
        & .061 & .047 & .643 & \textbf{.530} & .452 & .598
        & .069 & .055 & .650 & .573 & .326 & .446
        & .050 & .042 & .608 & .547 & .457 & .556
        & .067 & .056 & .647 & .589 & .355 & .437 \\

        EgoExo-WM (EgoX)
        & 6.1
        & .060 & .048 & \textbf{.633} & .541 & .465 & .554
        & .059 & .049 & .630 & .563 & .378 & .487
        & .050 & .041 & .610 & \textbf{.545} & .415 & .594
        & .063 & \textbf{.052} & \textbf{.640} & .581 & .395 & .456 \\

        \midrule

        \textbf{EgoExo-WM}
        & 6.1
        & \textbf{.058} & \textbf{.045} & \textbf{.633} & .535 & \textbf{.515} & \textbf{.618}
        & \textbf{.057} & \textbf{.047} & \textbf{.629} & \textbf{.561} & \textbf{.404} & \textbf{.531}
        & \textbf{.048} & \textbf{.040} & .611 & \textbf{.545} & \textbf{.489} & \textbf{.603}
        & \textbf{.062} & \textbf{.052} & \textbf{.640} & \textbf{.580} & \textbf{.460} & \textbf{.515} \\

        \bottomrule
    \end{tabular}
    \end{adjustbox}
    \caption{\textbf{Open-loop world model evaluation across four datasets.} 
    We report inference TFLOPs, final $2$s and average rollout $L_2$ error in DINOv3-L (DINO) and PE-Core-B (PE) space, and wrist accuracy. \textbf{\ourmethod} achieves the best overall performance, demonstrating the benefits of converted exocentric data. And, it does so with substantially lower inference compute than diffusion methods. Across $4$ datasets, \ourmethod's gains are statistically significant in $39$ of $42$ comparisons (video-level paired permutation tests, Holm-adjusted $p<0.05$; 
    see Appendix~\ref{app:rollout_statistical_sig}).
     $^*$ denotes adaptations: PEVA$^*$ predicts DINOv3-L latents rather than pixels, while EgoControl$^*$ matches Ego-WM's data and resolution. $^\dagger$ denotes the non-data-matched variant using $50$ hours of exocentric data ($240$ hours total). For diffusion baselines (PEVA$^*$, PEVA, EgoControl$^*$), we report means over 3 seeds, with standard errors $<.002$ (DINO), $<.006$ (PE), and $<.03$ (Wrist Acc).
    }
    \label{tab:wm_eval}
    \vspace{-1.75em}
\end{table*}

\noindent\textbf{Baselines and Ablations.}
We choose baselines to cover the key alternatives for prior egocentric world models that predict first-person futures from human motion. 
We further ablate our framework by holding the architecture and objective fixed while varying the use of exocentric data. Unless otherwise noted, all model variants use the same total training budget of $200$ hours and $224 \times 224$ resolution. PEVA, PEVA$^*$, and all our ablations use the L-sized architecture.

\begin{itemize}[leftmargin=*]
\item \noindent\textbf{PEVA} \citep{bai2025whole} is a SOTA diffusion-based egocentric video prediction model conditioned on human motion, trained on all $200$ hours of Nymeria with $7$ context frames.

\item \noindent\textbf{PEVA$^*$} adapts PEVA to predict DINOv3-L latents rather than pixels, controlling for the prediction space to assess whether our gains reflect better world modeling rather than an advantage from evaluation in latent space.

\item \noindent\textbf{EgoControl$^*$} \citep{pallotta2025egocontrol} is a SOTA human motion-conditioned egocentric video diffusion model fine-tuned from Cosmos-Predict2~\citep{nvidia2025cosmospredict2}. We adapt its 480p resolution to $224 \times 224$ and train on $200$ hours of Nymeria for consistency across methods. $^*$ denotes adapted implementation.


\item \noindent\textbf{EgoExo-WM (EgoX)} replaces \ouregoxspace with EgoX for exo-to-ego conversion while keeping the architecture and objective fixed, isolating the impact of pose-guided conversion on world model performance.

\item \noindent\textbf{Naive EgoExo-WM} uses the same architecture and objective, but trains on $190$ hours of Nymeria plus $10$ hours of raw exocentric video, isolating the benefit of exo-to-ego conversion. We also evaluate a variant with $50$ hours of raw exocentric video ($240$ hours total) to test whether more data can compensate for the lack of conversion.

\item \noindent\textbf{Ego-WM} is our egocentric-only variant, trained on $200$ hours of Nymeria with the same architecture and objective as our full model, but without exocentric data.

\end{itemize}

\noindent\textbf{EgoExo-WM} is our full method trained with $190$ hours of Nymeria and $10$ hours of converted exo-to-ego videos. 

\begin{table*}[t]
    \centering
    \scriptsize
    \setlength{\tabcolsep}{2.5pt}
    \renewcommand{\arraystretch}{1.12}
    \begin{adjustbox}{max width=\textwidth}
    \begin{tabular}{
        l
        *{3}{ccc@{\hspace{3pt}}}
        ccc@{}
    }
            \toprule
            \multirow{2}{*}{Model}
            & \multicolumn{3}{c}{\textbf{LEMMA}}
            & \multicolumn{3}{c}{\textbf{HOMAGE}}
            & \multicolumn{3}{c}{\textbf{Ego-Exo4D-Bike}}
            & \multicolumn{3}{c}{\textbf{Ego-Exo4D-Cooking}} \\
            \cmidrule(lr){2-4}
            \cmidrule(lr){5-7}
            \cmidrule(lr){8-10}
            \cmidrule(lr){11-13}
            & Acc $\uparrow$ & Body $\downarrow$ & Wrist $\downarrow$
            & Acc $\uparrow$ & Body $\downarrow$ & Wrist $\downarrow$
            & Acc $\uparrow$ & Body $\downarrow$ & Wrist $\downarrow$
            & Acc $\uparrow$ & Body $\downarrow$ & Wrist $\downarrow$ \\
            \midrule
            PEVA*
            & 30.1$\pm$0.9 & .484$\pm$.00 & .521$\pm$.00
            & 13.2$\pm$0.6 & .507$\pm$.01 & .566$\pm$.00
            & 33.2$\pm$1.1 & .290$\pm$.00 & .363$\pm$.00
            & 30.2$\pm$0.9 & .450$\pm$.00 & .508$\pm$.01 \\
            PEVA
            & 27.7$\pm$1.5 & .502$\pm$.02 & .547$\pm$.02
            & 8.9$\pm$0.4 & .506$\pm$.01 & .572$\pm$.01
            & 23.0$\pm$1.5 & .318$\pm$.01 & .406$\pm$.01
            & 21.7$\pm$0.4 & .468$\pm$.01 & .529$\pm$.01 \\
            EgoControl*
            & 29.9$\pm$0.0 & .476$\pm$.02 & .516$\pm$.02
            & \textbf{15.1$\pm$1.1} & \textbf{.459$\pm$.01} & \textbf{.520$\pm$.01}
            & 37.7$\pm$0.4 & .270$\pm$.00 & .342$\pm$.00
            & 31.8$\pm$1.9 & .445$\pm$.01 & .499$\pm$.02 \\
            \midrule
            Ego-WM
            & 35.3$\pm$0.9 & .422$\pm$.03 & .465$\pm$.03
            & 14.0$\pm$0.7 & .486$\pm$.02 & .545$\pm$.02
            & 37.8$\pm$1.2 & .272$\pm$.01 & .337$\pm$.01
            & 31.8$\pm$0.7 & .462$\pm$.00 & .517$\pm$.01 \\
            Naive EgoExo-WM
            & 33.8$\pm$0.4 & .445$\pm$.02 & .482$\pm$.02
            & 14.0$\pm$0.7 & .523$\pm$.01 & .579$\pm$.01
            & 39.3$\pm$0.6 & .264$\pm$.01 & .329$\pm$.01
            & 30.0$\pm$0.8 & .435$\pm$.02 & .493$\pm$.01 \\
            Naive EgoExo-WM$^\dagger$
            & 15.4$\pm$0.4 & .676$\pm$.01 & .732$\pm$.01
            & 12.3$\pm$0.4 & .524$\pm$.01 & .574$\pm$.01
            & 39.5$\pm$1.8 & .255$\pm$.01 & .317$\pm$.01
            & 30.2$\pm$0.6 & .436$\pm$.01 & .494$\pm$.01 \\
            EgoExo-WM (EgoX)
            & 31.6$\pm$1.7 & .469$\pm$.03 & .508$\pm$.03
            & 12.5$\pm$0.2 & .505$\pm$.00 & .563$\pm$.00
            & 38.5$\pm$1.5 & .260$\pm$.01 & .326$\pm$.01
            & 32.0$\pm$1.8 & .429$\pm$.02 & \textbf{.485$\pm$.02} \\
            \midrule
            EgoExo-WM
            & \textbf{37.7$\pm$1.1} & \textbf{.421$\pm$.02} & \textbf{.452$\pm$.02}
            & 13.0$\pm$0.8 & .510$\pm$.03 & .565$\pm$.03
            & \textbf{41.8$\pm$1.2} & \textbf{.248$\pm$.00} & \textbf{.313$\pm$.00}
            & \textbf{32.8$\pm$0.2} & \textbf{.428$\pm$.00} & .486$\pm$.01 \\
            \bottomrule
    \end{tabular}
    \end{adjustbox}
    \caption{
    \textbf{Model-Predictive Control Planning Results.}
    Acc measures ground-truth selection accuracy (\%), while Body and Wrist report MPJPE between selected and ground-truth motions. We report mean $\pm$ standard error over three candidate-sampling seeds, using identical action sequences across methods. Across four diverse datasets, \textbf{\ourmethodspace} achieves the strongest overall planning performance, highlighting its goal-directed action selection. $^*$ and $^\dagger$ as in Tab.~\ref{tab:wm_eval}.
    }
    \vspace{-1.75em}
    \label{tab:planning_eval}
\end{table*}

\vspace{-1.25em}
\paragraph{Evaluation Protocol.}
For open-loop world model evaluation, we use a $2$-second horizon ($8$ frames at $4$ Hz). \ourmethodspace and PEVA roll out autoregressively, feeding each prediction back with the next action. 
We report prediction error at the final rollout frame ($t=2s$) to assess long-horizon accuracy, and the average error across all $8$ predicted frames to assess overall rollout quality. 
Following prior latent world models, we evaluate in feature space to avoid introducing pixel reconstruction errors unrelated to prediction quality when decoding latent predictions, whereas pixel-based predictions can be directly encoded into the same feature space~\citep{baldassarre2025back, assran2025v, goswami2025world}. Specifically, we measure \textbf{visual prediction quality} using DINOv3-L $L_2$ distance, encoding pixel outputs with DINOv3-L and evaluating latent predictions directly. We measure \textbf{semantic prediction quality} using PE-Core-B~\citep{bolya2026perception} $L_2$ distance, encoding pixel outputs with PE-Core-B and mapping DINOv3 latents through a learned translator (Appendix~\ref{sec:pe_translation}).
Note that none of the models train for PE-Core-B. 
To complement the feature-space metrics, we evaluate \textbf{human-action consistency} using wrist accuracy (measured as PCK@$20$), with ViTPose~\citep{xu2022vitpose} for pixel outputs and our wrist keypoint predictor for latent predictions.


For model-predictive control planning, we evaluate each world model's ability to select goal-directed actions from a shared set of candidate trajectories. Following prior work~\citep{chen2026egoplan, chen2025worldprediction}, we construct $16$ candidate 3D human motion sequences of horizon $8$: the ground-truth sequence and $15$ alternatives sampled from other evaluation trajectories. This provides realistic human motion candidates while isolating the world model's ability to identify goal-reaching actions from proposal quality. Given the current observation and ground-truth future observation as the visual goal, each model rolls out the same candidates and selects the trajectory whose predicted final state is closest to the goal. We report selection accuracy (Acc), measuring the percentage of correctly selected ground-truth actions, alongside whole-body and wrist MPJPE, standard metrics for 3D human motion~\citep{hong2025egolm, ionescu2013human3, martinez2017human}. We report mean and standard error over $3$ candidate-sampling seeds, with all methods evaluated on the same action sequences for fair comparison.

Together, evaluation in DINOv3-L and PE-Core-B feature spaces captures complementary aspects of prediction quality, emphasizing spatial and semantic consistency, respectively, while avoiding pixel reconstruction artifacts. Wrist accuracy, planning accuracy, and MPJPE further assess embodied dynamics and planning utility. 

\begin{figure}[t]
    \centering
    \includegraphics[width=0.95\linewidth]{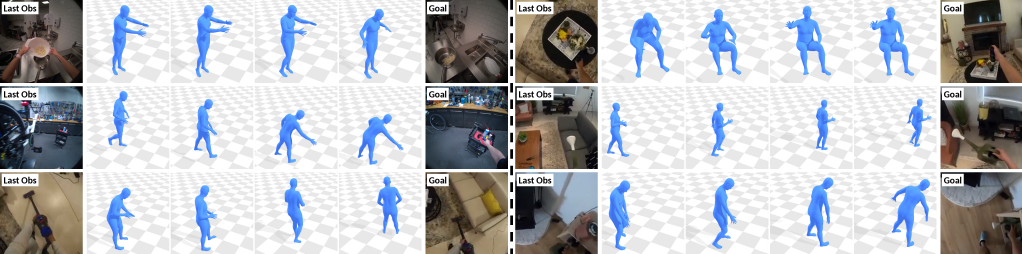}
    \caption{\textbf{Qualitative planning results.} 
    Given an observation and a visual goal, our world model ranks candidate motion sequences by their predicted proximity to the goal. Across activities including cooking, turning on a TV, bike repair, walking with a watering pot, and vacuuming, EgoExo-WM selects accurate motion. In the bottom-right failure case, it chooses an action sequence that steps forward and bends forward only slightly instead of bending down deeply in place. Videos are on our \projectpage{}.}
    \label{fig:wm_planning_qual}
    \vspace{-1.5em}
\end{figure}

\subsection{Direct Evaluation of World Model Accuracy}
\label{sec:rollout_eval}
Open-loop rollout evaluation is a standard approach for assessing whether a dynamics model accurately predicts future observations under a prescribed action sequence~\citep{bai2025whole, goswami2025world, pallotta2025egocontrol}. We evaluate \ourmethodspace against PEVA$^*$, PEVA and EgoControl$^*$, and conduct ablations to measure the contribution of our exocentric-to-egocentric training framework.

Table~\ref{tab:wm_eval} shows that \ourmethodspace outperforms PEVA$^*$, PEVA, and EgoControl$^*$ under a matched $200$-hour training budget, achieving lower embedding errors and higher wrist accuracy across all four datasets. Gains are particularly pronounced on HOMAGE and LEMMA---whose household and object-centric activities are underrepresented in Nymeria but unlocked in our converted exocentric data---with average DINO and PE $L_2$ errors reduced by 36--58\% and 11--24\%, respectively, relative to the baselines. This improvement is consistent with our method design: by converting diverse exocentric videos into egocentric training data, EgoExo-WM exposes the world model to a broader distribution of human motions, object interactions, and environments. Improvements over PEVA$^*$, which predicts DINOv3-L latents, show that these gains are not merely due to the prediction space. Furthermore, \ourmethodspace uses at least \textbf{$\mathbf{58\times}$ less inference compute} than baselines.

We ablate \ourmethodspace against variants using EgoX instead of \ouregoxspace (\ourmethodspace (EgoX)), omitting exo-to-ego conversion (Naive \ourmethod, including the $50$-hour raw exocentric variant), or excluding exocentric data (Ego-WM). \ourmethodspace performs best overall across four datasets, again with the largest gains on HOMAGE and LEMMA, further demonstrating the benefits of our unlocked converted exocentric data. 
PE and DINO $L_2$ gains are smaller on Bike and Cooking, the latter potentially due to strong cooking coverage in Nymeria. \ourmethodspace consistently improves wrist accuracy, indicating better fine-grained action consistency. Gains over the EgoX-based variant in DINO $L_2$ and wrist accuracy highlight how \ouregoxspace better preserves visual dynamics and human-motion consistency, while smaller PE $L_2$ gains reflect similar semantic coverage from shared exocentric data.
Together, these comparisons show that gains come not simply from adding exocentric data, but from effectively converting it into egocentric, action-aligned trajectories suitable for world-model training.

\subsection{Impact for Model-Predictive Control Planning}
\label{sec:wm_planning_results}
Next we demonstrate that \ourmethodspace can be used for goal-directed planning. The model can evaluate candidate human motion action sequences and select the one most likely to reach a desired visual goal. This requires reasoning about both navigation and coordinated body motion: for example, reaching for a mug may require walking to the shelf, tiptoeing, and reaching. Our experiment isolates whether \ourmethodspace improves whole-body action selection compared to alternative world-model evaluators.

Table~\ref{tab:planning_eval} shows that \ourmethodspace achieves the strongest overall planning performance among PEVA$^*$, PEVA, EgoControl$^*$, and our world-model baselines, obtaining the best results on eight of twelve metrics. \ourmethodspace achieves the best results on LEMMA, Ego-Exo4D-Bike, and Ego-Exo4D-Cooking, except for Cooking wrist MPJPE, where it is within $0.001$ of the best. HOMAGE is the most challenging planning dataset, with the lowest accuracy across all methods. \ourmethodspace sometimes selects trajectories with similar body motions but incorrect displacement. Gains over Naive \ourmethod, which actually \emph{degrades} when scaling raw exocentric data from $10$ to $50$ hours, show that naive scaling is insufficient. These gains reflect the benefit of incorporating diverse exocentric videos converted into action-aligned egocentric observations, providing broader coverage of environments, body motions, and object-centric actions than egocentric data alone. This allows \ourmethodspace to better evaluate candidate actions, improving whole-body planning across datasets. See Fig.~\ref{fig:wm_planning_qual} and our \projectpage{} for more qualitative planning examples.

\section{Limitations and Conclusions}
Despite our progress, there are a few known limitations that call for continued research. 
\ourmethodspace focuses on short-term interactions; extending exo-to-ego conversion to longer activities and planning under compounding prediction errors remain future work. Conversion also struggles with complex human-object interactions, severe occlusions, and small-object manipulation. 

In conclusion, we presented \ourmethod, a framework for training egocentric world models from exocentric video by leveraging recovered 3D human motion to guide exo-to-ego conversion and provide action conditioning. Our results show that converted exocentric data improves egocentric prediction and goal-directed planning across diverse datasets, suggesting a scalable path toward learning egocentric models from abundant exocentric human video. Ultimately, this work takes a step toward enabling embodied agents to learn from observing others and use that knowledge to plan their own interactions with the world.

\section{LLM Usage Statement}
We used ChatGPT throughout the manuscript to refine original drafts, improve clarity and organization, and assist with grammar and editing. Claude Code was similarly used to assist in creating and refining figures based on provided designs and ideas. Claude Code and Codex were also used to assist with code implementation and execution of experiments, including model training, rollout evaluation, and planning. The authors verified all experiments, results, and text.

\newpage

{
    \small
    \bibliographystyle{plainnat}
    \bibliography{main}
}

\medskip


\newpage

\appendix

\section*{Technical appendices and supplementary material}

%
%
%

\begin{figure*}[t]
    \centering
    \includegraphics[width=\textwidth]{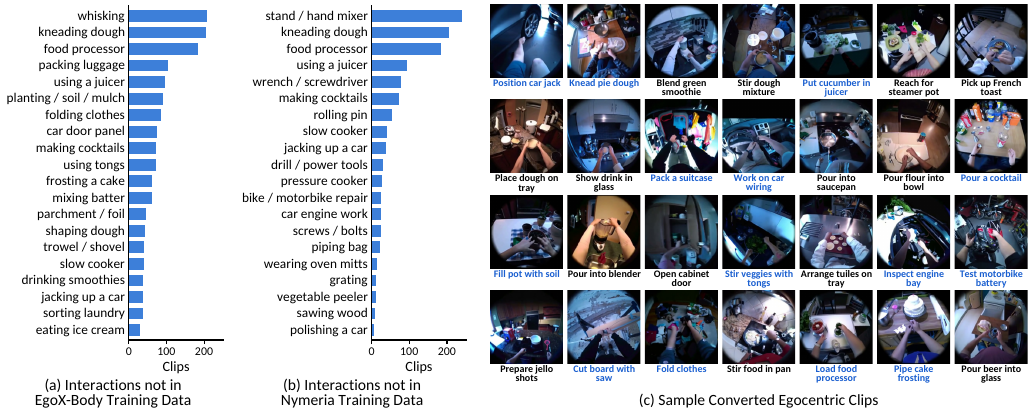}
    \caption{\textbf{Converted training data.} (a) The 20 most frequent interactions in the converted clips that appear in none of \ouregox's training clips. (b) The 20 most frequent interactions that appear in none of the 642 narrated recordings in our Nymeria training split, the egocentric data that Ego-WM trains on. (c) Sample egocentric frames from converted clips. Blue labels mark clips that show an interaction absent from either dataset. See Sec.~\ref{sec:training_data_stats} for details.}
    \label{fig:converted_stats}
\end{figure*}

\suppressfloats[t] 
In this appendix, we provide full ablation results, additional analyses, and implementation details. We first present manual evaluation protocols, component ablations, sensitivity analyses, and additional qualitative results for \ouregoxspace (Sec.~\ref{sec:egox_analysis}). We then provide our statistical analysis for our rollout evaluation (Sec.~\ref{app:rollout_statistical_sig}). We then provide details about our DINOv3 to PE-Core-B feature translator (Sec.~\ref{sec:pe_translation}). We then describe exocentric video selection, filtering, post-processing, and training-data statistics, including the interactions the converted clips contain beyond \ouregox's training data and beyond Nymeria (Sec.~\ref{sec:dataset_details} and Fig.~\ref{fig:converted_stats}). Finally, we provide implementation details for the world model and \ouregox, including architectures and training procedures (Sec.~\ref{sec:impl_details}).

\section{\ouregoxspace Analysis}
\label{sec:egox_analysis}

\subsection{Manual Evaluation Protocol}
\label{sec:manual_eval}

\begin{wraptable}{R}{0.45\textwidth}
    \centering
    \vspace{-4mm}   
    {%
    \scriptsize
    \setlength{\tabcolsep}{3pt}
    \renewcommand{\arraystretch}{0.95}
    \begin{adjustbox}{max width=\linewidth}
    \begin{tabular}{llrrr}
        \toprule
        Analysis & Bin & Ego-Exo4D & 100DOH & Total \\
        \midrule
        Pose estimator & All & 100 & 99 & 199 \\
        Component ablation & All & 100 & 200 & 300 \\
        \midrule
        \multirow{4}{*}{Action complexity}
            & Low & 29 & 28 & 57 \\
            & Medium & 38 & 38 & 76 \\
            & High & 33 & 33 & 66 \\
            & \textbf{Total} & \textbf{100} & \textbf{99} & \textbf{199} \\
        \midrule
        \multirow{5}{*}{Body occlusion}
            & Clear & 25 & 25 & 50 \\
            & Mild & 25 & 25 & 50 \\
            & Moderate & 25 & 25 & 50 \\
            & Heavy & 25 & 25 & 50 \\
            & \textbf{Total} & \textbf{100} & \textbf{100} & \textbf{200} \\
        \midrule
        \multirow{5}{*}{Viewpoint}
            & Front & 25 & 25 & 50 \\
            & Side & 25 & 25 & 50 \\
            & Back & 25 & 25 & 50 \\
            & Top-down & 25 & 25 & 50 \\
            & \textbf{Total} & \textbf{100} & \textbf{100} & \textbf{200} \\
        \bottomrule
    \end{tabular}
    \end{adjustbox}
    \par}
    \captionsetup{position=bottom, skip=7pt}
    \caption{Sample counts for manual conversion evaluation.}
    \label{tab:manual_eval_samples}
    \vspace{-7mm}   
\end{wraptable}
\paragraph{Evaluation criteria.}
We conduct a manual evaluation to analyze the sensitivity of EgoX-Body to different model components and input video characteristics. A single evaluator familiar with the task assessed whether each generated video correctly performed the intended action. A conversion was considered unsuccessful if it exhibited flipped arms, incorrect arm motion, or incorrect object interactions. Figs.~\ref{fig:supp_egox_vs_body_1} and~\ref{fig:supp_egox_vs_body_2} (Sec.~\ref{sec:egox_more_qual}) show examples of each failure type in EgoX conversions: flipped arms (Fig.~\ref{fig:supp_egox_vs_body_1}, bottom row), incorrect arm motion (Fig.~\ref{fig:supp_egox_vs_body_1}, top left), and incorrect object interactions (Fig.~\ref{fig:supp_egox_vs_body_1}, top right, and Fig.~\ref{fig:supp_egox_vs_body_2}). Figs.~\ref{fig:supp_more_egox_body_1} and~\ref{fig:supp_more_egox_body_2} show \ouregoxspace conversions without these failures. The same failure criteria were applied across all analyses.

\paragraph{Automated stratification.}
For the pose-estimator sensitivity, component ablation, viewpoint, body-occlusion, and action-complexity analyses, the evaluator assessed conversion failures without access to the clips' assigned analysis bins. In particular, the viewpoint, body-occlusion, and action-complexity categories were determined automatically rather than manually annotated.

\begin{itemize}
    \item \textbf{Viewpoint.} We combine CLIP-based front, side, back, and overhead viewpoint evidence with pose geometry, including left--right shoulder and hip ordering and body foreshortening. Clips are ranked automatically, and 25 unique, activity-diverse clips are selected for each viewpoint category in each dataset.

    \item \textbf{Body occlusion.} Pose visibility is estimated at frames 0, 12, 24, 36, and 48, with missing detections assigned zero visibility. Eligible clips must contain a person detection in at least three of the five sampled frames and have a mean detection confidence of at least 0.75. Dataset-specific visibility quartiles define the clear, mild, moderate, and heavy occlusion categories.

    \item \textbf{Action complexity.} Clips are categorized according to the number of distinct captioned actions: low (0--2), medium (3--4), and high (5 or more).
\end{itemize}

\paragraph{Evaluation samples.}
Table~\ref{tab:manual_eval_samples} summarizes the number of evaluated clips for each analysis across Ego-Exo4D and 100DOH.

\paragraph{Limitations and scope.}
Since the evaluation was conducted by a single evaluator, inter-annotator agreement could not be measured. However, failure judgments were performed without knowledge of the analysis bins, and viewpoint, visibility, and action-complexity categories were determined independently through automated procedures. These analyses are intended to characterize conversion behavior under different model configurations and input conditions.

\subsection{Component Ablation}
\label{sec:component_ablation}

\begin{wraptable}{R}{0.45\textwidth}
    \centering
    \vspace{-4mm}   
    {%
    \scriptsize
    \setlength{\tabcolsep}{3pt}
    \renewcommand{\arraystretch}{1.05}
    \begin{adjustbox}{max width=\linewidth}
    \begin{tabular}{lccc}
        \toprule
        Method & Flip $\downarrow$ &
        \shortstack{Incorrect motion/\\interaction $\downarrow$} &
        \shortstack{Any\\failure $\downarrow$} \\
        \midrule
        \multicolumn{4}{l}{\textbf{100DOH}} \\
        \textbf{EgoX-Body} & \textbf{1.5} & \textbf{4.5} & \textbf{6.0} \\
        w/o body overlay & 2.0 & 6.5 & 8.5 \\
        w/o hand prior & 7.0 & 23.0 & 27.5 \\
        \midrule
        \multicolumn{4}{l}{\textbf{Ego-Exo4D}} \\
        \textbf{EgoX-Body} & \textbf{0.0} & \textbf{37.0} & \textbf{37.0} \\
        w/o body overlay & 1.0 & 39.0 & 40.0 \\
        w/o hand prior & 1.0 & 55.0 & 55.0 \\
        \bottomrule
    \end{tabular}
    \end{adjustbox}
    \par}
    \captionsetup{position=bottom, skip=7pt}
    \caption{Component ablation on 100DOH and Ego-Exo4D. We report conversion failure rates (\%, lower is better).}
    \label{tab:component_ablation}
    \vspace{-5mm}   
\end{wraptable}
We examine the contributions of the body skeleton overlay and egocentric hand prior to EgoX-Body's conversion quality in Table~\ref{tab:component_ablation}. We evaluate two types of failures: \emph{flipped arms}, where the generated video reverses the intended arm action, and \emph{incorrect motion/interaction}, where the generated arm motion or object interaction does not match the intended action.

Removing the egocentric hand prior leads to substantially larger increases in conversion failures than removing the body overlay across both datasets. The hand prior provides guidance on hand location, handedness, and object-relative movement, while the body overlay supplies complementary global-pose guidance. These results suggest that the egocentric hand prior plays a more significant role in maintaining accurate arm motion and object interactions.

\subsection{Viewpoint Sensitivity}
\label{sec:viewpoint}

\begin{wraptable}{R}{0.45\textwidth}
    \centering
    \vspace{-4mm}   
    {%
    \scriptsize
    \setlength{\tabcolsep}{5pt}
    \renewcommand{\arraystretch}{1.05}
    \begin{adjustbox}{max width=\linewidth}
    \begin{tabular}{lcc}
        \toprule
        Viewpoint & Ego-Exo4D $\downarrow$ & 100DOH $\downarrow$ \\
        \midrule
        \textbf{Front} & \textbf{28.0} & \textbf{12.0} \\
        Side & 32.0 & 29.0 \\
        Back & 36.0 & 36.0 \\
        Top-down & 52.0 & 16.0 \\
        \bottomrule
    \end{tabular}
    \end{adjustbox}
    \par}
    \captionsetup{position=bottom, skip=7pt}
    \caption{Conversion failure rates (\%) across viewpoints on Ego-Exo4D and 100DOH. Lower is better.}
    \label{tab:viewpoint}
    \vspace{-5mm}   
\end{wraptable}
We analyze the effect of viewpoint on conversion quality (Table~\ref{tab:viewpoint}). Viewpoint is automatically determined by combining CLIP-based front, side, back, and overhead viewpoint evidence with five-frame pose geometry, including left--right body ordering, shoulder/hip foreshortening, and head-to-hip foreshortening. We evaluate 25 clips per viewpoint for each dataset.

Front views yield the lowest conversion failure rates on both datasets, likely because they provide clearer visibility of the body and interaction workspace. Side and back views exhibit higher failure rates, potentially due to increased self-occlusion and object occlusion. Top-down views show dataset-dependent performance, with substantially higher failure rates on Ego-Exo4D, where the body may be incomplete or strongly foreshortened. Front views have one distinct failure mode: when the actor faces the camera, the space in front of them is not visible in the exocentric video, so the generated scene can be wrong (Fig.~\ref{fig:supp_egox_body_limitations}, top left).

\subsection{Body Occlusion}
\label{sec:body_occlusion}

\begin{wraptable}{R}{0.45\textwidth}
    \centering
    \vspace{-4mm}   
    {%
    \scriptsize
    \setlength{\tabcolsep}{5pt}
    \renewcommand{\arraystretch}{1.05}
    \begin{adjustbox}{max width=\linewidth}
    \begin{tabular}{lcc}
        \toprule
        Visibility & Ego-Exo4D $\downarrow$ & 100DOH $\downarrow$ \\
        \midrule
        \textbf{Clear} & \textbf{20.0} & \textbf{16.0} \\
        Mild & 24.0 & 20.0 \\
        Moderate & 64.0 & 24.0 \\
        Heavy & 40.0 & 33.0 \\
        \bottomrule
    \end{tabular}
    \end{adjustbox}
    \par}
    \captionsetup{position=bottom, skip=7pt}
    \caption{Conversion failure rates (\%) across body occlusion levels on Ego-Exo4D and 100DOH. Lower is better.}
    \label{tab:body_occlusion}
    \vspace{-5mm}   
\end{wraptable}
We analyze the effect of body occlusion on conversion quality (Table~\ref{tab:body_occlusion}). We estimate pose visibility by averaging calibrated Keypoint R-CNN confidence scores over 12 COCO torso and limb joints across five sampled frames, with missing detections assigned zero visibility. Occlusion is defined as one minus visibility. Clips are grouped into clear, mild, moderate, and heavy occlusion categories using dataset-specific quartiles, with 25 clips per category for each dataset.

On 100DOH, conversion failure rates increase consistently with occlusion severity. Ego-Exo4D exhibits a non-monotonic trend, but clear and mildly occluded clips remain substantially more reliable than moderately and heavily occluded clips. These results suggest that mild occlusion is generally tolerated, while stronger occlusion reduces the body, hand, and interaction cues needed for accurate conversion.

\subsection{Action Complexity}
\label{sec:action_complexity}

\begin{wraptable}{R}{0.45\textwidth}
    \centering
    \vspace{-4mm}   
    {%
    \scriptsize
    \setlength{\tabcolsep}{5pt}
    \renewcommand{\arraystretch}{1.05}
    \begin{adjustbox}{max width=\linewidth}
    \begin{tabular}{lcc}
        \toprule
        Complexity & Ego-Exo4D $\downarrow$ & 100DOH $\downarrow$ \\
        \midrule
        Low & 34.5 & 17.9 \\
        Medium & 50.0 & \textbf{15.8} \\
        High & \textbf{27.3} & 36.4 \\
        \bottomrule
    \end{tabular}
    \end{adjustbox}
    \par}
    \captionsetup{position=bottom, skip=7pt}
    \caption{Conversion failure rates (\%) across action complexity levels on Ego-Exo4D and 100DOH. Lower is better.}
    \label{tab:action_complexity}
    \vspace{-5mm}   
\end{wraptable}
We analyze the effect of action complexity on conversion quality (Table~\ref{tab:action_complexity}). Complexity is automatically determined by counting the number of distinct actions described in each clip's caption. Clips are categorized as low (0--2 actions), medium (3--4 actions), or high (5 or more actions).

On 100DOH, clips with high action complexity exhibit substantially higher failure rates than those with low or medium complexity. The trend is less consistent on Ego-Exo4D, where high-complexity clips have the lowest failure rate. This suggests that action complexity alone does not determine conversion difficulty, which may also depend on viewpoint, visibility, and the nature of the interactions.

\subsection{Pose-Estimator Sensitivity}
\label{sec:pose_estimator_sensitivity}

\begin{wraptable}{R}{0.45\textwidth}
    \centering
    \vspace{-4mm}   
    {%
    \scriptsize
    \setlength{\tabcolsep}{5pt}
    \renewcommand{\arraystretch}{1.05}
    \begin{adjustbox}{max width=\linewidth}
    \begin{tabular}{lcc}
        \toprule
        Pose estimator & Ego-Exo4D $\downarrow$ & 100DOH $\downarrow$ \\
        \midrule
        SMPLer-X S32 & 46.0 & 33.0 \\
        SMPLer-X H32 & 46.0 & 28.3 \\
        \textbf{SAM-Body4D} & \textbf{38.0} & \textbf{23.2} \\
        \bottomrule
    \end{tabular}
    \end{adjustbox}
    \par}
    \captionsetup{position=bottom, skip=7pt}
    \caption{Conversion failure rates (\%) using different pose estimators on Ego-Exo4D and 100DOH. Lower is better.}
    \label{tab:pose_estimator_sensitivity}
    \vspace{-5mm}   
\end{wraptable}
Our source clips consist of opportunistic real-world videos with self-occlusion, object occlusion, motion blur, and imperfect viewpoints. We adopt SAM-Body4D for body pose estimation during inference and HaMeR for hand pose estimation during training, as preliminary qualitative comparisons found them to produce more accurate and temporally consistent pose estimates than alternatives. We evaluate the sensitivity of EgoX-Body to the choice of pose estimator by replacing SAM-Body4D with SMPLer-X (S32 and H32 variants; Table~\ref{tab:pose_estimator_sensitivity}).

SAM-Body4D achieves the lowest conversion failure rates on both datasets, outperforming both SMPLer-X variants. These results indicate that conversion quality is sensitive to the underlying pose estimator, particularly for challenging real-world videos with occlusion and imperfect viewpoints. Nevertheless, SMPLer-X still produces successful conversions for the majority of clips (54\% on Ego-Exo4D and 67--72\% on 100DOH), suggesting that our conversion pipeline remains reasonably robust to the choice of pose estimator.

\subsection{Additional Qualitative Results}
\label{sec:egox_more_qual}

\begin{figure*}[p]
    \centering
    \includegraphics[width=\textwidth]{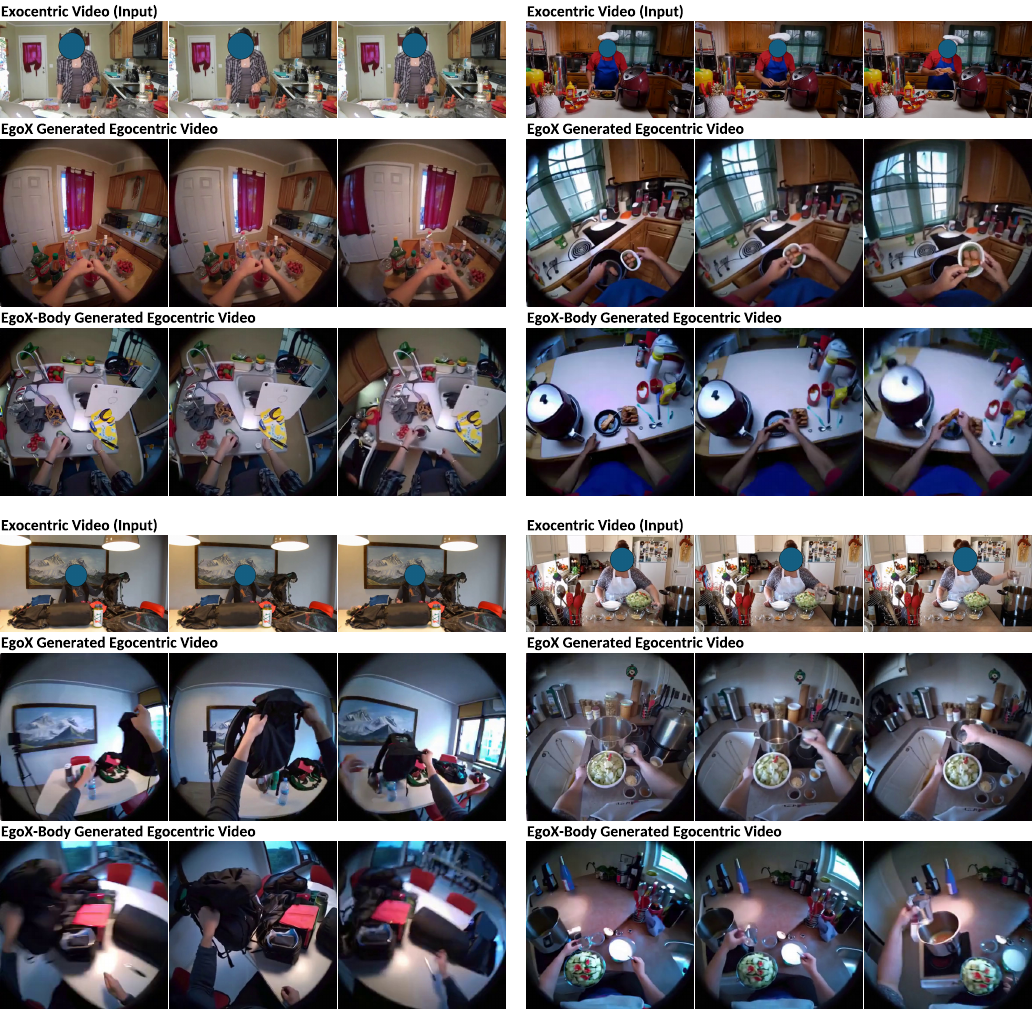}
    \caption{\textbf{Additional EgoX vs.\ \ouregoxspace comparisons (1/2).} For each example, we show the exocentric input video (top) and the egocentric videos generated by EgoX (middle) and \ouregoxspace (bottom). \emph{Top left:} EgoX raises both arms, inconsistent with the exocentric video, and renders the wrong scene, whereas \ouregox's arm movements are consistent with the exocentric video. \emph{Top right:} EgoX's hand--object interaction is inconsistent and its scene is wrong, whereas \ouregoxspace produces consistent arm movements in the correct scene. \emph{Bottom left:} EgoX swaps the arms, holding the bag with the right arm, whereas \ouregox's arm movements are consistent. \emph{Bottom right:} EgoX swaps the arms, grabbing the glass with the right arm, whereas \ouregox's arm movements are consistent.}
    \label{fig:supp_egox_vs_body_1}
\end{figure*}

\begin{figure*}[t]
    \centering
    \includegraphics[width=\textwidth]{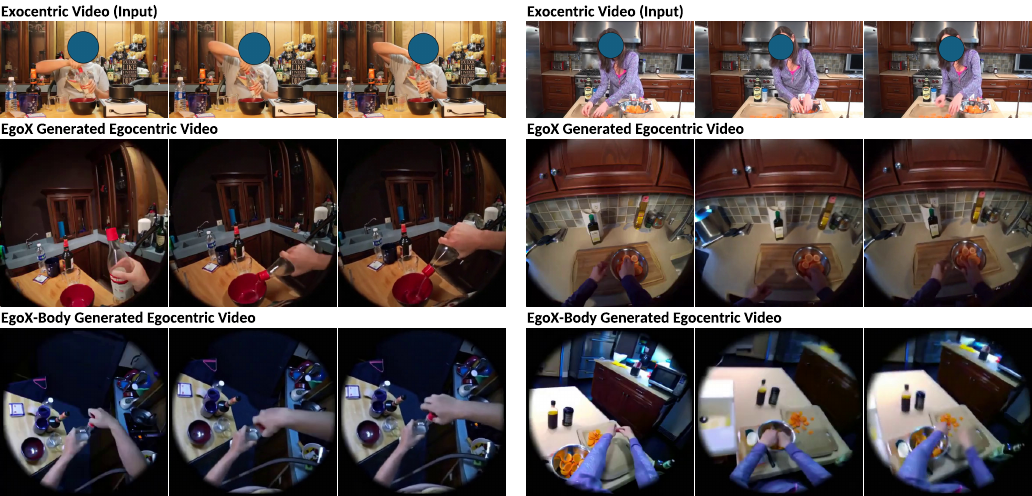}
    \caption{\textbf{Additional EgoX vs.\ \ouregoxspace comparisons (2/2).} Layout as in Fig.~\ref{fig:supp_egox_vs_body_1}. \emph{Left:} EgoX's hand--object interaction is inconsistent with the input, whereas \ouregox's is relatively consistent, although it struggles with opening the bottle. \emph{Right:} EgoX's hand--object interaction is inconsistent, whereas \ouregox's is relatively consistent.}
    \label{fig:supp_egox_vs_body_2}
\end{figure*}

\begin{figure*}[t]
    \centering
    \includegraphics[width=\textwidth]{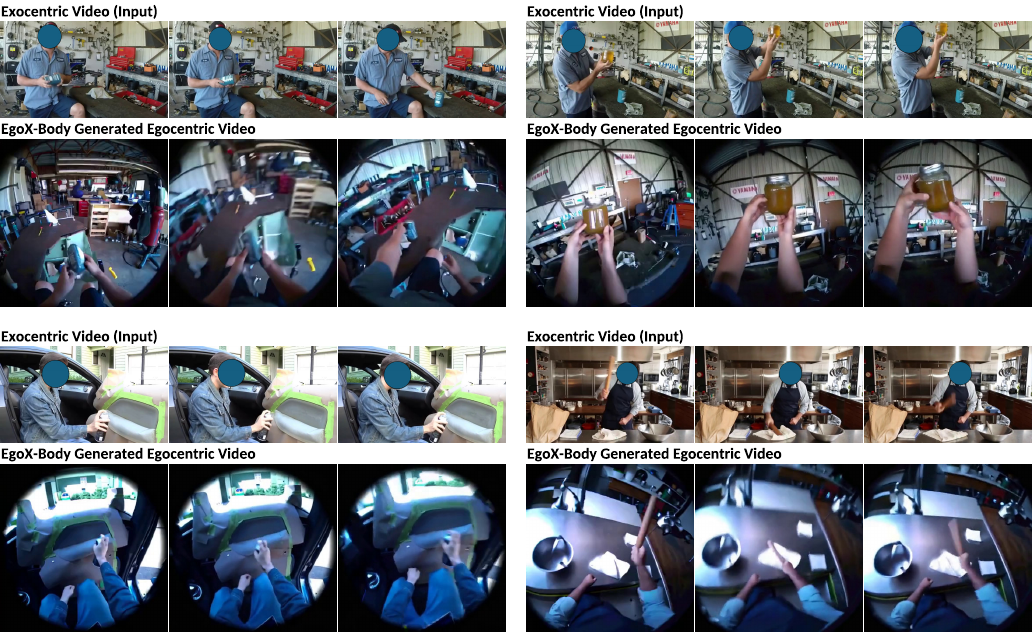}
    \caption{\textbf{Additional \ouregoxspace results (1/2).} For each example, we show the exocentric input video (top) and the egocentric video generated by \ouregoxspace (bottom).}
    \label{fig:supp_more_egox_body_1}
\end{figure*}

\begin{figure*}[t]
    \centering
    \includegraphics[width=\textwidth]{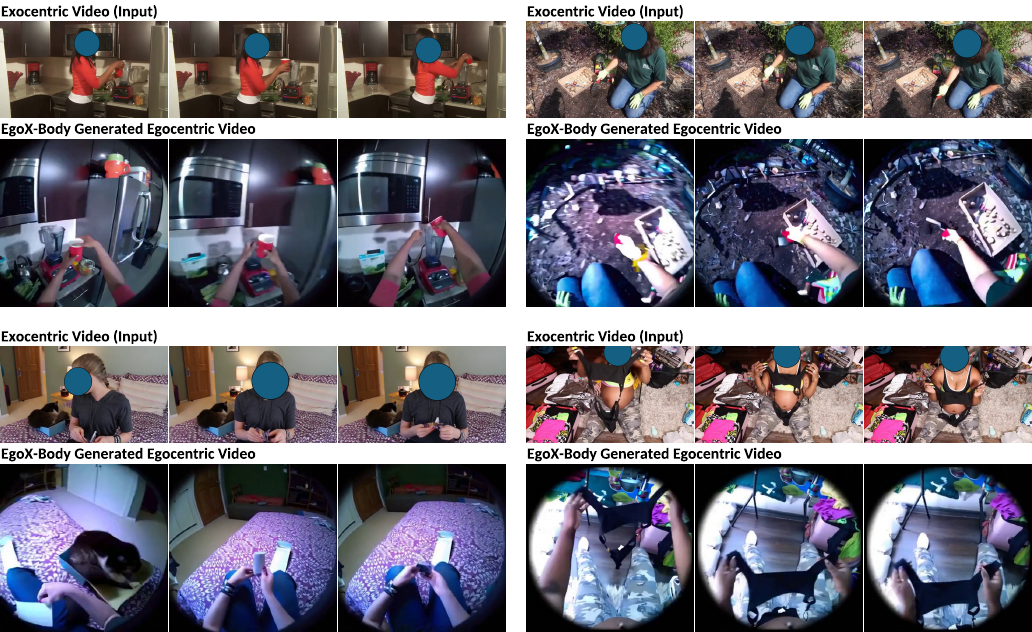}
    \caption{\textbf{Additional \ouregoxspace results (2/2).} Layout as in Fig.~\ref{fig:supp_more_egox_body_1}.}
    \label{fig:supp_more_egox_body_2}
\end{figure*}

\begin{figure*}[t]
    \centering
    \includegraphics[width=\textwidth]{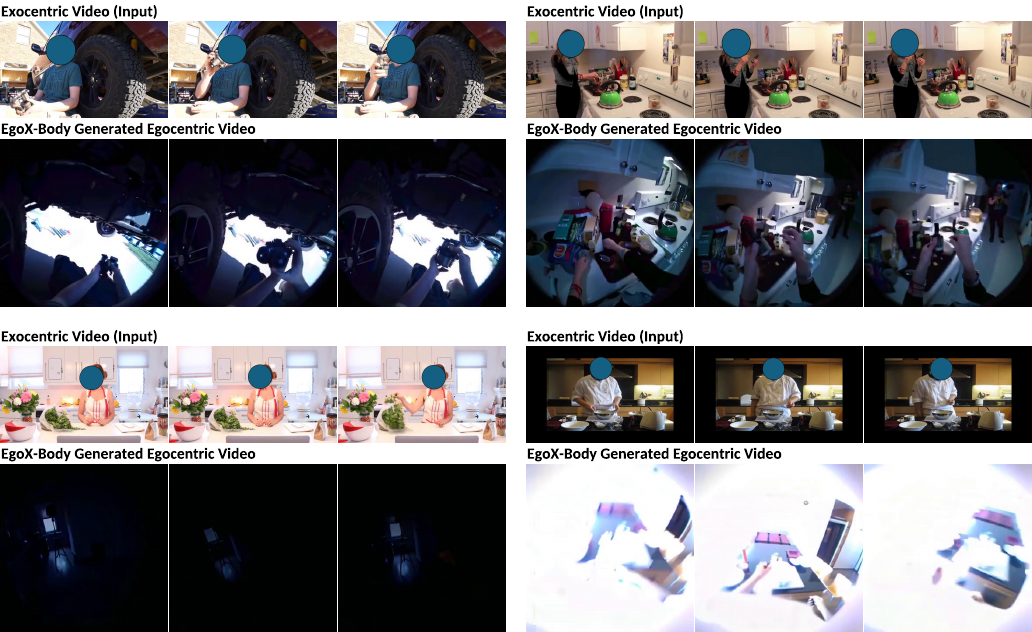}
    \caption{\textbf{\ouregoxspace failure modes.} \emph{Top left:} when the actor faces the camera, there is no egocentric prior and the generated scene context is wrong. \emph{Top right:} small objects can appear out of nowhere. \emph{Bottom left:} black prediction. \emph{Bottom right:} oversaturated (white) prediction. Our automatic quality filter removes nearly black or oversaturated clips (Sec.~\ref{app:filtering-criteria}).}
    \label{fig:supp_egox_body_limitations}
\end{figure*}

We provide additional examples in the style of Fig.~\ref{fig:egox_body_qualitative}, each showing the first, middle, and last frames of a 3-second clip. Figs.~\ref{fig:supp_egox_vs_body_1} and~\ref{fig:supp_egox_vs_body_2} compare EgoX and \ouregoxspace on the same exocentric inputs. EgoX often swaps the left and right arms, generates hand--object interactions that do not match the input, or renders the wrong scene, whereas \ouregoxspace follows the actor's arm motion. Figs.~\ref{fig:supp_more_egox_body_1} and~\ref{fig:supp_more_egox_body_2} show further \ouregoxspace conversions across workshop, kitchen, car, garden, and bedroom scenes. Many of these interactions never appear in \ouregox's training data (Sec.~\ref{sec:training_data_stats} and Fig.~\ref{fig:converted_stats}(a)). Fig.~\ref{fig:supp_egox_body_limitations} shows typical failure modes of \ouregox. When the actor faces the camera, the scene in front of them is not visible in the exocentric video, so the egocentric prior is nearly empty and the generated scene is wrong. Small objects can also appear without support in the input, and some generations are nearly black or oversaturated (see also Fig.~\ref{fig:egox-failures}), which our automatic quality filter removes before world model training (Sec.~\ref{app:filtering-criteria}).

\section{Rollout Evaluation Statistical Significance Test}
\label{app:rollout_statistical_sig}

\paragraph{Method.}
We compare EgoExo-WM against seven baselines across six metrics, using source videos as independent units to account for correlated clips (81, 92, 49, and 81 videos across LEMMA, HOMAGE, Bike, and Cooking). For each comparison, we compute the signed improvement, averaged equally across datasets. We use 20{,}000 paired permutations with video-level sign flips and Holm correction over 42 comparisons. We obtain 95\% confidence intervals from 20{,}000 video-level bootstrap resamples within each dataset. For stochastic baselines, we average per-clip scores over three seeds before testing. Regenerated per-clip scores closely match Table~\ref{tab:wm_eval}.

\paragraph{Results.}
Table~\ref{tab:wm_significance} shows that EgoExo-WM significantly outperforms baselines in 39/42 comparisons (Holm-adjusted $p<.05$), including all 18 against diffusion-based models. The exceptions are PE at both horizons against EgoExo-WM (EgoX) and 2s wrist accuracy against Ego-WM. The former reflects similar semantic coverage between \ourmethodspace and \ourmethodspace (EgoX) since they convert the same exocentric data. 

\begin{table*}[t]
    \centering
    \footnotesize
    \setlength{\tabcolsep}{5pt}
    \newcommand{\sigval}[1]{{\boldmath$#1$}}
    \begin{adjustbox}{max width=\textwidth}
    \begin{tabular}{lcccccc}
        \toprule
        \multirow{2}{*}{EgoExo-WM vs.}
        & \multicolumn{2}{c}{DINO ($10^{-3}$)}
        & \multicolumn{2}{c}{PE ($10^{-3}$)}
        & \multicolumn{2}{c}{Wrist Acc (points)} \\
        \cmidrule(lr){2-3}
        \cmidrule(lr){4-5}
        \cmidrule(lr){6-7}
        & 2s & Avg & 2s & Avg & 2s & Avg \\
        \midrule
        PEVA$^*$
        & \begin{tabular}[c]{@{}c@{}}\sigval{+27.7} \\ {\tiny$[26.5, 29.0]$}\end{tabular}
        & \begin{tabular}[c]{@{}c@{}}\sigval{+24.4} \\ {\tiny$[23.5, 25.3]$}\end{tabular}
        & \begin{tabular}[c]{@{}c@{}}\sigval{+37.6} \\ {\tiny$[32.3, 43.0]$}\end{tabular}
        & \begin{tabular}[c]{@{}c@{}}\sigval{+52.0} \\ {\tiny$[47.4, 56.8]$}\end{tabular}
        & \begin{tabular}[c]{@{}c@{}}\sigval{+16.4} \\ {\tiny$[11.5, 21.2]$}\end{tabular}
        & \begin{tabular}[c]{@{}c@{}}\sigval{+18.6} \\ {\tiny$[15.6, 21.4]$}\end{tabular} \\
        PEVA
        & \begin{tabular}[c]{@{}c@{}}\sigval{+55.1} \\ {\tiny$[53.8, 56.3]$}\end{tabular}
        & \begin{tabular}[c]{@{}c@{}}\sigval{+56.9} \\ {\tiny$[55.8, 57.9]$}\end{tabular}
        & \begin{tabular}[c]{@{}c@{}}\sigval{+100.8} \\ {\tiny$[95.3, 106.4]$}\end{tabular}
        & \begin{tabular}[c]{@{}c@{}}\sigval{+137.3} \\ {\tiny$[132.8, 141.7]$}\end{tabular}
        & \begin{tabular}[c]{@{}c@{}}\sigval{+17.0} \\ {\tiny$[11.2, 22.5]$}\end{tabular}
        & \begin{tabular}[c]{@{}c@{}}\sigval{+17.4} \\ {\tiny$[14.4, 20.3]$}\end{tabular} \\
        EgoControl$^*$
        & \begin{tabular}[c]{@{}c@{}}\sigval{+35.4} \\ {\tiny$[33.4, 37.3]$}\end{tabular}
        & \begin{tabular}[c]{@{}c@{}}\sigval{+32.7} \\ {\tiny$[30.7, 34.6]$}\end{tabular}
        & \begin{tabular}[c]{@{}c@{}}\sigval{+63.1} \\ {\tiny$[55.4, 70.6]$}\end{tabular}
        & \begin{tabular}[c]{@{}c@{}}\sigval{+75.6} \\ {\tiny$[67.0, 83.8]$}\end{tabular}
        & \begin{tabular}[c]{@{}c@{}}\sigval{+13.3} \\ {\tiny$[7.7, 18.9]$}\end{tabular}
        & \begin{tabular}[c]{@{}c@{}}\sigval{+10.9} \\ {\tiny$[7.8, 14.0]$}\end{tabular} \\
        \midrule
        Ego-WM
        & \begin{tabular}[c]{@{}c@{}}\sigval{+6.4} \\ {\tiny$[5.7, 7.0]$}\end{tabular}
        & \begin{tabular}[c]{@{}c@{}}\sigval{+6.2} \\ {\tiny$[5.7, 6.6]$}\end{tabular}
        & \begin{tabular}[c]{@{}c@{}}\sigval{+7.8} \\ {\tiny$[4.0, 11.6]$}\end{tabular}
        & \begin{tabular}[c]{@{}c@{}}\sigval{+19.6} \\ {\tiny$[17.1, 22.1]$}\end{tabular}
        & \begin{tabular}[c]{@{}c@{}}$+5.3$ \\ {\tiny$[0.3, 10.5]$}\end{tabular}
        & \begin{tabular}[c]{@{}c@{}}\sigval{+7.9} \\ {\tiny$[5.1, 10.6]$}\end{tabular} \\
        Naive EgoExo-WM
        & \begin{tabular}[c]{@{}c@{}}\sigval{+3.4} \\ {\tiny$[2.7, 4.1]$}\end{tabular}
        & \begin{tabular}[c]{@{}c@{}}\sigval{+2.6} \\ {\tiny$[2.1, 3.1]$}\end{tabular}
        & \begin{tabular}[c]{@{}c@{}}\sigval{+9.7} \\ {\tiny$[5.4, 13.8]$}\end{tabular}
        & \begin{tabular}[c]{@{}c@{}}\sigval{+11.1} \\ {\tiny$[8.2, 13.9]$}\end{tabular}
        & \begin{tabular}[c]{@{}c@{}}\sigval{+9.6} \\ {\tiny$[4.1, 15.2]$}\end{tabular}
        & \begin{tabular}[c]{@{}c@{}}\sigval{+7.5} \\ {\tiny$[5.0, 9.9]$}\end{tabular} \\
        Naive EgoExo-WM (50h Exo)$^\dagger$
        & \begin{tabular}[c]{@{}c@{}}\sigval{+4.8} \\ {\tiny$[4.0, 5.6]$}\end{tabular}
        & \begin{tabular}[c]{@{}c@{}}\sigval{+3.4} \\ {\tiny$[2.9, 4.0]$}\end{tabular}
        & \begin{tabular}[c]{@{}c@{}}\sigval{+8.6} \\ {\tiny$[3.8, 13.3]$}\end{tabular}
        & \begin{tabular}[c]{@{}c@{}}\sigval{+4.3} \\ {\tiny$[1.2, 7.5]$}\end{tabular}
        & \begin{tabular}[c]{@{}c@{}}\sigval{+7.0} \\ {\tiny$[1.9, 12.1]$}\end{tabular}
        & \begin{tabular}[c]{@{}c@{}}\sigval{+5.6} \\ {\tiny$[3.2, 8.0]$}\end{tabular} \\
        EgoExo-WM (EgoX)
        & \begin{tabular}[c]{@{}c@{}}\sigval{+1.2} \\ {\tiny$[0.6, 1.9]$}\end{tabular}
        & \begin{tabular}[c]{@{}c@{}}\sigval{+1.1} \\ {\tiny$[0.7, 1.5]$}\end{tabular}
        & \begin{tabular}[c]{@{}c@{}}$-0.4$ \\ {\tiny$[-4.1, 3.3]$}\end{tabular}
        & \begin{tabular}[c]{@{}c@{}}$+2.2$ \\ {\tiny$[-0.2, 4.5]$}\end{tabular}
        & \begin{tabular}[c]{@{}c@{}}\sigval{+6.6} \\ {\tiny$[2.1, 11.1]$}\end{tabular}
        & \begin{tabular}[c]{@{}c@{}}\sigval{+4.7} \\ {\tiny$[2.3, 7.0]$}\end{tabular} \\
        \bottomrule
    \end{tabular}
    \end{adjustbox}
    \caption{\textbf{Statistical significance averaged across four datasets.} EgoExo-WM improvements over each baseline (positive is better), with 95\% video-level bootstrap confidence intervals. DINO and PE differences are in $10^{-3}$; wrist accuracy in percentage points. \textbf{Bold} denotes statistical significance (Holm-adjusted $p<.05$ over 42 tests). Confidence intervals are unadjusted.}
    \label{tab:wm_significance}
\end{table*}

\section{DINOv3 to PE-Core-B Feature Translator}
\label{sec:pe_translation}
To evaluate DINOv3-based world models in the PE-Core-B feature space, we train a lightweight transformer to map DINOv3-L patch tokens to the patch tokens of PE-Core-B16-224~\cite{bolya2026perception}. The translator is trained only on features extracted from real video frames of Ego4D, Ego-Exo4D, LEMMA, and HOMAGE, excluding the evaluation videos; it never sees world-model predictions or rollouts.

\paragraph{Training Data and Architecture.}
We sample approximately 500K Ego4D frames at least 15 seconds apart, excluding videos overlapping with downstream evaluation sets, and split them by participant into training and 22,200 validation frames. We add approximately 84K unlabeled frames from Ego-Exo4D, LEMMA, and HOMAGE, including the evaluation videos, for 534,528 training frames. Each frame is resized to $224 \times 224$ and encoded by frozen DINOv3-L into $196 \times 1024$ patch tokens. The target is the frame's final-layer $14 \times 14$ patch tokens from frozen PE-Core-B/16 at its native $224 \times 224$ resolution, taken before attention pooling. These align one-to-one with the DINOv3 grid, giving $196 \times 768$ target tokens, each $\ell_2$-normalized.

The translator consists of a LayerNorm and linear projection to 768 dimensions, learned positional embeddings, 12 pre-norm Transformer layers with 12 attention heads, and a token-wise MLP ($768 \rightarrow 2048 \rightarrow 768$). All 196 spatial tokens are preserved, and the output tokens are $\ell_2$-normalized. The model contains approximately 89M parameters and is trained from scratch.

\paragraph{Optimization and Model Selection.}
We optimize a per-token cosine loss and a relational loss ($\lambda = 30$) that preserves the teacher's frame-to-frame similarities within each batch. Each batch contains 256 frames: 128 sampled frames, each paired with an adjacent frame from the same video. We train for 6 epochs with AdamW, using a learning rate of $10^{-4}$ ($3\times10^{-4}$ for the output MLP), weight decay of 0.05, 5\% warmup followed by cosine decay, and gradient clipping at 1.0. The translator never sees actions, poses, or world-model outputs.

\section{Dataset Details}
\label{sec:dataset_details}

\subsection{Exocentric Video Selection and Filtering}
\label{sec:exo_filtering}
To construct valid training clips from raw videos, we apply an automatic scene filtering pipeline. Each video is first partitioned into candidate scenes using adaptive scene detection, and scenes below a minimum duration are removed. For each remaining scene, we sample three representative frames and require a single visible person to be detected in all of them. This is checked using ViTPose~\citep{xu2022vitpose}, where a scene is retained only when reliable head and upper-body keypoints are present. We also estimate camera motion over the scene using ORB correspondences and affine motion estimation, rejecting clips with noticeable zoom, translation, or rotation.

The remaining candidates are passed to GPT-4o-mini as a conservative visual validator. Given the sampled frames, GPT-4o-mini~\citep{achiam2023gpt} predicts whether the scene contains a clear human action, estimates the maximum overlay coverage, and identifies photographic overlays such as picture-in-picture videos. We accept a scene only when a visible human action is present, the estimated overlay coverage is below $20\%$, and no photographic overlay is detected. Fig.~\ref{fig:vlm-filtering-prompt} shows the prompt we use.

\paragraph{VLM Filtering Prompt.}
\begin{figure*}[t]
    \centering
    \begin{minipage}{0.95\textwidth}
    \footnotesize
\begin{verbatim}
Check these images and answer STRICTLY in JSON.

You are judging the scene across all provided images together.

Definitions:
- A "human action" means a visible person actively doing something.
- human_action should be true ONLY if a person is clearly performing
  a visible action in at least one image, and the scene overall appears
  to depict a human action.
  Examples: cutting, cooking, typing, walking, opening something,
  holding or manipulating objects, cleaning, assembling, playing.

- human_action should be false for:
  - static poses (standing, sitting without action)
  - portraits/selfies
  - empty scenes
  - landscapes
  - objects moving without human involvement
  - animals acting without humans
  - a person just talking to the camera (e.g. vlog-style, speaking, presenting)
  - a person facing the camera without clear interaction with objects

Overlay rules:
- An "overlay" is anything drawn on top of the main scene.
- overlay_is_photographic should be true ONLY if any image contains a
  natural photo or video image as an overlay (e.g. picture-in-picture, inset video).
- overlay_is_photographic should be false for logos, branding, text,
  subtitles, UI elements, icons, flat illustrations, watermarks.

Rules:
- human_action: true only if a clear human action is visible across the set
- overlay_pct: estimated percent (0-100) of the image covered by overlays;
  if images differ, use the highest / worst-case estimate
- passes = human_action AND overlay_pct < 20 AND (overlay_is_photographic == false)

Be conservative if unsure.
\end{verbatim}
    \end{minipage}
    \caption{\textbf{VLM filtering prompt.} Exact prompt used
    to filter exocentric video clips.}
    \label{fig:vlm-filtering-prompt}
\end{figure*}

\subsubsection{Post-Processing}
\label{app:postprocessing}

EgoX-Body converts each exocentric clip into a $49$-frame egocentric sequence at $16$ Hz. For world model training, we temporally downsample these converted clips to $8$ Hz, yielding $25$ frames per sequence. We apply the same temporal sampling to the corresponding 3D human-motion trajectories, so each converted clip becomes an egocentric observation-action sequence aligned with the format of real egocentric trajectories. This allows converted exocentric interactions to be used jointly with real egocentric data during training.

After conversion, we standardize generated clips for world model training. Each converted video is saved at $384 \times 384$ resolution and $16$ Hz. We center crop at $85\%$, resize to $224 \times 224$, and then we downsample to $8$ Hz for world model training and pair each egocentric observation sequence with the corresponding whole-body action trajectory derived from the estimated body pose.

\subsubsection{Failure Cases}
\label{app:failure-cases}

\begin{figure}[t]
    \centering
    \includegraphics[width=\linewidth]{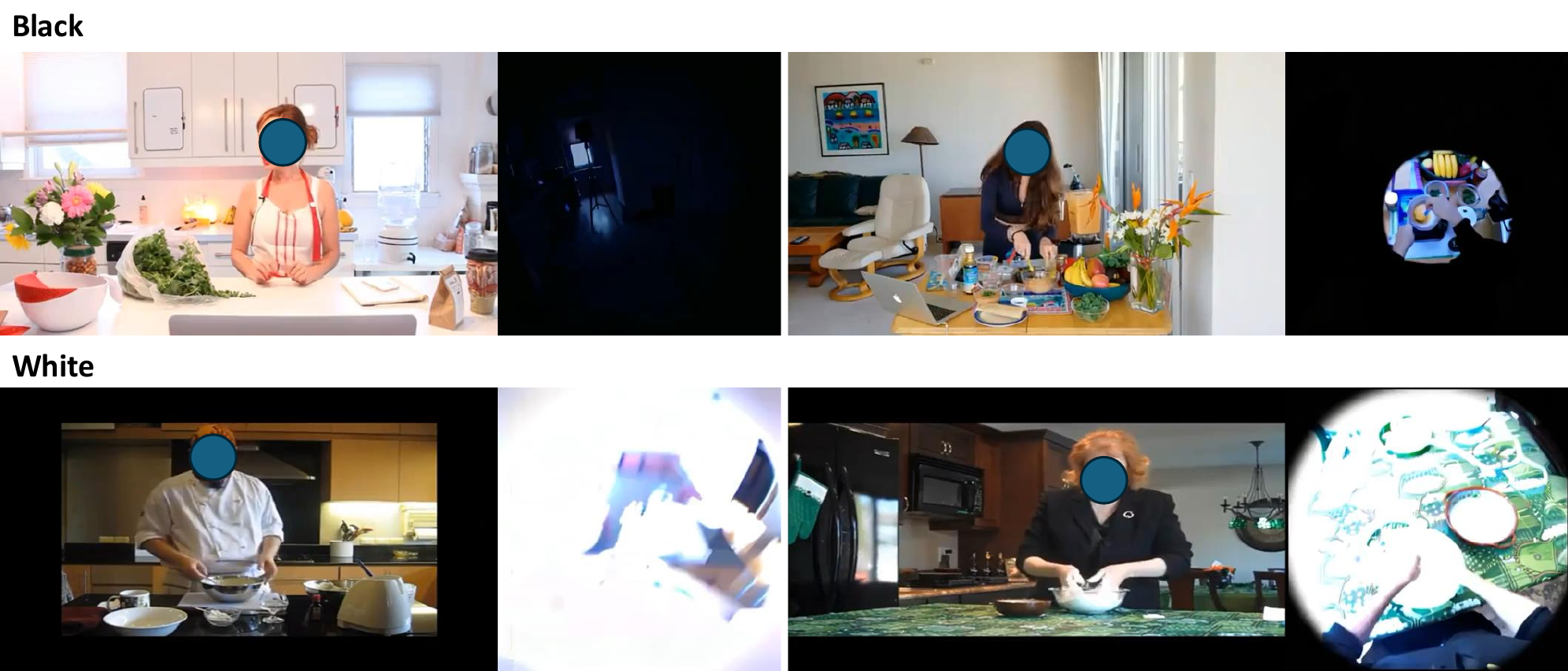}
    \caption{\textbf{Examples of failure cases in Internet ego-view videos.} We show representative clips with large white or black regions, which commonly arise from videos where the person is directly facing the camera where the egocentric prior is essentially black. These failure cases provide little useful training signal, motivating the automatic filtering criteria described in Section~\ref{app:filtering-criteria}.}
    \label{fig:egox-failures}
\end{figure}

We observe that Internet ego-view collections contain several recurring failure cases that make them unhelpful for training. These include black or near-black cuts, title cards or overexposed frames, motion-blurred clips, and sped-up tutorials with abrupt frame-to-frame changes. Such clips either lack useful visual content or contain temporal artifacts that can degrade world model training. We provide some qualitative examples of failure cases in Figure~\ref{fig:egox-failures} and in the videos on our \projectpage{}.

\subsubsection{Filtering EgoX-Body Generated Videos}
\label{app:filtering-criteria}

To remove these failure cases, we implement a simple automatic quality filter on the $326 \times 326$ ego crop of each source clip. Specifically, we compute four scalar statistics: \texttt{black\_fraction\_mean}, the average fraction of near-black pixels over time; \texttt{white\_fraction\_mean}, the average fraction of near-white pixels over time; \texttt{blur\_median}, the median Laplacian variance across frames; and \texttt{motion\_median}, the median mean-absolute-difference between consecutive grayscale frames. We retain clips satisfying \texttt{black\_fraction\_mean} $< 0.30$, \texttt{white\_fraction\_mean} $< 0.20$, \texttt{blur\_median} $> 50$, and \texttt{motion\_median} $< 32.5$. These thresholds are calibrated by inspecting random clips near each cutoff to verify that they separate clear failure cases from usable footage. This filtering retains roughly $80\%$ of candidate clips for training.

\subsection{Training-Data Statistics}
\label{sec:training_data_stats}

\begin{wraptable}{R}{0.45\textwidth}
    \centering
    \vspace{-4mm}   
    {%
    \scriptsize
    \setlength{\tabcolsep}{5pt}
    \renewcommand{\arraystretch}{1.05}
    \begin{adjustbox}{max width=\linewidth}
    \begin{tabular}{lrc}
        \toprule
        Source & Clips & Share (\%) \\
        \midrule
        100DOH~\citep{shan2020understanding} & 4,972 & 43.8 \\
        HowTo100M~\citep{miech2019howto100m} & 5,321 & 46.9 \\
        CrossTask~\citep{zhukov2019cross} & 1,048 & 9.2 \\
        \midrule
        \textbf{Total} & \textbf{11,341} & \textbf{100.0} \\
        \bottomrule
    \end{tabular}
    \end{adjustbox}
    \par}
    \captionsetup{position=bottom, skip=7pt}
    \caption{Distribution of converted training clips across source datasets.}
    \label{tab:training_data_stats}
    \vspace{-0.8mm}   
\end{wraptable}
Our converted training set contains 11,341 clips from 650 source videos, spanning three datasets. Table~\ref{tab:training_data_stats} summarizes the distribution of clips across these sources.

\textbf{Action and object distributions.}
We estimate action and object distributions from video captions using multi-label categorization. Common actions include holding/grasping (72.6\%), moving/transferring (57.7\%), stabilizing (45.2\%), adjusting (23.4\%), lifting (22.2\%), stirring (21.7\%), and pouring (17.4\%). Frequently occurring object categories include containers (53.3\%), food (43.5\%), work surfaces (39.2\%), utensils (25.2\%), and appliances (13.1\%).

\textbf{Activity domains.}
The converted data is predominantly cooking (58.6\%) and beverage preparation (18.8\%), but it also covers packing (6.9\%), household chores and cleaning (6.4\%), repair (3.6\%), furniture and DIY (2.7\%), gardening (1.7\%), and vehicle maintenance (1.2\%).

\textbf{Comparison with \ouregox's training data.}
We compare the coverage of our converted clips, \ouregox's paired training data (3,818 clips from Ego-Exo4D and H2O), and Nymeria using their captions or narrations. Fig.~\ref{fig:converted_stats}(a) shows the 20 most frequent interactions absent from \ouregox's training data. These include whisking (207 clips), kneading dough (206), packing luggage (103), gardening (92), folding clothes (85), refinishing car door panels (76), and frosting cakes (62). \ouregoxspace also converts clips from unseen environments, including household (464 clips) and vehicle settings (354). More broadly, \ouregox's training data covers only Ego-Exo4D's cooking, sports, dance, bike repair, health, and music scenarios and H2O's tabletop scenes, with no gardening, furniture and DIY work (including woodworking), vehicle maintenance, packing, or household chores. Despite limited paired ego-exo training data, \ouregoxspace generalizes to diverse unseen interactions and environments, as illustrated in Figs.~\ref{fig:egox_body_qualitative}, \ref{fig:converted_stats}(c), \ref{fig:supp_egox_vs_body_1}, \ref{fig:supp_egox_vs_body_2}, \ref{fig:supp_more_egox_body_1}, and~\ref{fig:supp_more_egox_body_2}, as well as in the videos on our \projectpage{}.

\textbf{Comparison with Nymeria world model training data.}
Nymeria constitutes all of Ego-WM's training data and the egocentric portion of EgoExo-WM's training data. We verify matching Nymeria narrations to exclude object mentions and mimed actions. We identify 28 interactions absent from Nymeria. These include using diverse tool use, repair activities, using mixers, kneading dough, using food processors (Fig.~\ref{fig:converted_stats}(b) and Fig.~\ref{fig:converted_stats}(c)). While these counts are caption-derived estimates with potentially overlapping categories, they demonstrate how exocentric video expands the diversity of interactions available for world model training.

\section{Implementation Details}
\label{sec:impl_details}

\subsection{World Model Implementation Details}
\label{app:wm-details}

\subsubsection{Architecture and Training Details}
We train our world model with a CDiT-L/2~\citep{bar2025navigation, bai2025whole} backbone consisting of $24$ layers, hidden size $1024$, $16$ attention heads, and MLP ratio $4$. The model takes as input DINOv3 ViT-L/16 patch tokens extracted from $224{\times}224$ images, producing a $14{\times}14$ token grid with $196$ tokens of dimension $1024$. We condition on $3$.

We optimize with AdamW using a constant learning rate of $8{\times}10^{-5}$, $\beta_1{=}0.9$, $\beta_2{=}0.95$, zero weight decay, gradient clipping at $10.0$, and \texttt{bfloat16} mixed precision. We train with a per-GPU batch size of $64$ on $8$ A40 GPUs, resulting in an effective global batch size of $512$ and train for a total of $100,000$ iterations. For evaluation, we use an exponential moving average of the model weights, and the model is compiled with \texttt{torch.compile} during training and evaluation.

\subsubsection{Action Representation}
\label{app:smpl-action-space}
We use a 3D human-motion action space based on the $22$-joint SMPL body skeleton, as described in the main paper in Section~\ref{sec:wm_training}. PEVA and EgoControl use Xsens motion trajectories with $23$ body joints, while our model uses the $22$-joint SMPL skeleton. We therefore convert actions to the skeleton required by each model. The primary mismatch is the torso chain: Xsens uses four spine joints, \texttt{L5}, \texttt{L3}, \texttt{T12}, and \texttt{T8}, while SMPL uses three. For Xsens-to-SMPL conversion, we drop the intermediate lumbar joint \texttt{L3} and map \texttt{L5}, \texttt{T12}, and \texttt{T8} to the three SMPL spine joints. For SMPL-to-Xsens conversion, we duplicate the corresponding lumbar spine joint to recover the \texttt{L3} entry. This conversion only changes the skeleton layout; the action parameterization remains unchanged.

\subsubsection{Wrist-Position Consistency Objective}
\label{app:wrist-consistency}
The wrist consistency loss encourages the world model's predicted next-frame representations to remain spatially faithful to hand placement. We attach a frozen wrist-heatmap decoder on top of the world model's predicted DINOv3 patch tokens and penalize disagreement with a 2D ground-truth heatmap. The decoder is a 6-layer Transformer head (256-dim, 16 heads, MLP ratio 4) that takes the $14{\times}14$ grid of DINOv3 ViT-L/16 patch tokens (dim 1024) from a $224{\times}224$ image and predicts a $16{\times}16$ local patch per token, tiled back into a single-channel $224{\times}224$ heatmap. The decoder is pretrained on EgoExo4D and Nymeria with per-pixel MSE against ground-truth Gaussian heatmaps before being frozen for world-model training. 

Ground-truth heatmaps are generated from ViTPose wrist keypoints, keeping only detections with mean confidence above $0.3$ and de-duplicating wrists within $5$~px of each other. Both wrists are rendered onto a single channel by splatting an isotropic 2D Gaussian with $\sigma = 3.0$~px and combining the two via element-wise max, which preserves a unit peak when the wrists are co-located. During world-model training, the frozen decoder is applied to the predicted next-frame patch tokens, and we add an MSE loss between the resulting heatmap and the ground-truth, masked to frames in which at least one wrist is visible.

\subsection{\ouregoxspace Details}
\label{app:egox-body-details}

\subsubsection{Release}
\label{sec:egox_release}
When releasing EgoX-Body, which entails image generation, we will state guidelines for use.

\subsubsection{Compute Details}
We train \ouregoxspace on 4 GH$200$s for $20,000$ iterations and perform inference on GH$200$ gpus with each sample taking of $49$ frames taking $3.25$ minutes. 

\subsubsection{Inference Runtime}
\label{sec:egox_body_inference_runtime}
\begin{wraptable}{R}{0.45\textwidth}
    \centering
    \vspace{-4mm}   
    {%
    \begin{adjustbox}{max width=\linewidth}
    \begin{tabular}{lc}
        \toprule
        \textbf{Stage} & \textbf{Mean Time} \\
        \midrule
        ViPE (depth/camera) & 20.0 s \\
        SAM (masks) & 4.7 s \\
        SAM-Body4D (inference) & 4.3 s \\
        EgoPrior (rendering) & 8.5 s \\
        EgoX-Body (generation) & 3 min 15 s \\
        \midrule
        \textbf{Total} & \textbf{3 min 52.5 s} \\
        \bottomrule
    \end{tabular}
    \end{adjustbox}
    \par}
    \captionsetup{position=bottom, skip=7pt}
    \caption{Average preprocessing time per video for each stage of the exo-to-ego conversion pipeline.}
    \label{tab:egox_body_inference_runtime}
    \vspace{-10mm}   
\end{wraptable}
Table~\ref{tab:egox_body_inference_runtime} reports the average runtime of each stage in our exo-to-ego conversion pipeline. This is measured on an NVIDIA GH200 GPU. The full pipeline takes approximately 3.9 minutes per video.

\subsubsection{Body Pose and Scene Reconstruction}
\label{app:body-scene-reconstruction}

Given an in-the-wild exocentric video, we estimate full-body pose using SAM-Body4D and obtain a 4D scene reconstruction using ViPE~\citep{huang2025vipe} following EgoX. The extracted body pose is converted from the MHR representation to SMPL-X for compatibility with the rest of the pipeline. The body pose provides the actor's motion and 3D hand positions, while the 4D reconstruction provides scene geometry for rendering an approximate egocentric view.

\subsubsection{Egocentric Prior and Hand Overlay}
\label{app:ego-prior-hand-overlay}

We render the SMPL-X skeleton recovered from the exocentric video into an egocentric view using a pinhole projection from a head-anchored virtual camera. The virtual camera center is defined as the midpoint of the SMPL-X eye joints, \texttt{L\_EYE}=23 and \texttt{R\_EYE}=24. We additionally push the virtual camera forward $0.1$m and remove points too close to avoid rendering artifacts. We construct the camera basis using the eye-to-eye direction as the horizontal axis $\mathbf{x}$, the camera-center-to-\texttt{NECK} direction as the vertical axis $\mathbf{y}$, and $\mathbf{z}=\mathbf{x}\times\mathbf{y}$ as the forward-looking axis.

For the hand overlay, we use the SMPL-X wrist and finger joints for each hand, with $16$ joints and $15$ bones per hand. Bones connect the wrist to each finger and proceed distally along the kinematic chain. The left hand is drawn in red and the right hand in blue, with circles for joints and lines for bones.

\subsubsection{Practical Modifications}
\label{app:egox-body-practical-modifications}
We make several practical modifications to improve robustness and efficiency. We add $250$ training samples from the H2o~\citep{kwon2021h2o} dataset, focusing on sequences where the person faces the camera, which better matches many in-the-wild exocentric videos. We remove the Geometry Guided Attention (GGA) module from EgoX, reduce the generation resolution from $448 \times 448$ to $384 \times 384$, and save generated videos at $16$ Hz. On an NVIDIA GH200, these changes reduce the video generation inference time from approximately $17.5$ minutes per video to approximately $3.25$ minutes per video. 


\end{document}